\documentclass[compsoc,conference,a4paper,10pt,times]{IEEEtran}
\IEEEoverridecommandlockouts
\usepackage[nocompress]{cite}
\usepackage{amsmath,amssymb,amsfonts}
\usepackage{algorithmic}
\usepackage{graphicx}
\usepackage{textcomp}
\usepackage{bmpsize}
\usepackage[dvipsnames]{xcolor}
\usepackage{lipsum}
\usepackage{hyperref}
\hypersetup{colorlinks,urlcolor=black}
\def\BibTeX{{\rm B\kern-.05em{\sc i\kern-.025em b}\kern-.08em
    T\kern-.1667em\lower.7ex\hbox{E}\kern-.125emX}}

\usepackage[small]{caption}
\usepackage{graphicx}
\usepackage{amsbsy}
\usepackage{bbm}
\usepackage{booktabs}
\usepackage{dsfont}
\usepackage[]{footmisc}
\usepackage{listings}
\usepackage{multirow}
\usepackage{multicol}
\usepackage{makecell}
\usepackage{subcaption}
\usepackage{xr}
\usepackage{wasysym} 
\usepackage{enumitem}

\newcommand{\name}{\textsc{Illuminati}}

\newcommand{\revision}{\color{black}{}}

\begin{document}


\title{{\name}: Towards Explaining Graph Neural Networks for \\ Cybersecurity Analysis}

\author{\IEEEauthorblockN{Haoyu He}
\IEEEauthorblockA{
The George Washington University\\
haoyuhe@gwu.edu}
\and
\IEEEauthorblockN{Yuede Ji}
\IEEEauthorblockA{University of North Texas\\
yuede.ji@unt.edu}
\and
\IEEEauthorblockN{H. Howie Huang}
\IEEEauthorblockA{
The George Washington University\\
howie@gwu.edu}
}

\maketitle


\begin{abstract}


Graph neural networks (GNNs) have been utilized to create multi-layer graph models for a number of cybersecurity applications from fraud detection to software vulnerability analysis.
Unfortunately, like traditional neural networks, GNNs also suffer from a lack of transparency, that is, it is challenging to interpret the model predictions. Prior works focused on specific factor explanations for a GNN model. 
In this work, we have designed and implemented {\name}, a comprehensive and accurate explanation framework for cybersecurity applications using GNN models. Given a graph and a pre-trained GNN model, {\name} is able to identify the important nodes, edges, and attributes that are contributing to the prediction while requiring no prior knowledge of GNN models. We evaluate {\name} in two cybersecurity applications, i.e., code vulnerability detection and smart contract vulnerability detection. The experiments show that {\name} achieves more accurate explanation results than state-of-the-art methods, specifically, 87.6\% of subgraphs identified by {\name} are able to retain their original prediction, an improvement of 10.3\% over others at 77.3\%. 
Furthermore, the explanation of {\name} can be easily understood by the domain experts, suggesting the significant usefulness for the development of cybersecurity applications.






\end{abstract}

\section{Introduction}
\label{sec_introduction}


{\revision
Graph is a structured data representation with nodes and edges, where nodes denote the entities and edges denote the relationship between them.
Graph has been widely used in cybersecurity applications, such as code property graph for code vulnerability detection~\cite{yamaguchi2014modeling}, API-call graph for Android malware detection~\cite{xu2019droidevolver}, and website network for malicious website detection~\cite{li2013ieee}. 
}

Graph neural networks (GNNs) are multi-layer neural networks that can learn representative embeddings on structured graph data~\cite{gnn2019}. 
Because of that, GNNs have achieved outstanding performance for various cybersecurity applications, such as malicious account detection~\cite{liu2018heterogeneous,wang2019fdgars}, fraud detection~\cite{liu2020alleviating, dou2020enhancing}, software vulnerability detection~\cite{devign, cheng2021deepwukong, cao2021bgnn4vd}, memory forensic analysis~\cite{song2018deepmem}, and binary code analysis~\cite{xu2017neural, li2019graph, ji2021buggraph}. 
{\revision
Existing works usually construct graphs from an application and train a GNN model that can learn the node or graph representation. The GNN model can be used for various downstream tasks, e.g., node classification~\cite{kipf2017semi}, link prediction~\cite{zhang2018link}, and graph classification~\cite{Errica2020A}.
Taking binary code similarity detection as an example, recent works~\cite{xu2017neural,ji2021buggraph} first transform binary code into an attributed control flow graph. With that graph, they train a GNN model that can represent each graph as an embedding. Finally, they use a similarity function, e.g., cosine similarity, to measure code similarity. 
}

{\revision
\subsection{Motivation}
When a pre-trained GNN model is deployed in reality, it usually generates many positive alarms that need to be manually verified by the cybersecurity analysts to confirm their existence. Unfortunately, existing models usually generate too many alarms that the cybersecurity analysts are not able to verify them in a timely manner, which is known as the threat alert fatigue problem~\cite{hassan2019nodoze}. According to a recent study from FireEye, most organizations in US receive 17,000 alters per week while only 4\% of them are properly investigated~\cite{fireeye}.

To investigate a generated alarm, the cybersecurity analysts usually need to manually figure out why it is predicted as a positive. If such information can be provided automatically, it would greatly help to accelerate the manual investigation process. 
Unfortunately, GNN models lack the explainability similar to traditional deep neural networks.
There have been efforts towards automatically explaining neural networks, such as convolutional neural networks~\cite{Zhang_2018_CVPR}, recurrent neural networks~\cite{bradbury2016quasi}. 
However, they cannot be directly applied because GNNs work on the graph, which is an irregular data structure. 
Each node in a graph can have arbitrary neighbors and the order may be arbitrary as well. 
Therefore, the traditional explanation methods would fail to explain the interaction between node attributes without considering the message passing through edges.

On the other hand, several GNN explanation methods are proposed recently~\cite{pgmexplainer,subgraphx_icml21,luo2020parameterized,GNNEx19,schlichtkrull2021interpreting}. However, these methods mainly aim to provide an explanation of certain factors from the input graphs.
Table~\ref{tb:explainers} compares the recent works for GNN explanation.
In particular, PGM-Explainer~\cite{pgmexplainer} and SubgraphX~\cite{subgraphx_icml21} apply a \textit{node-centric} strategy to identify the important nodes as the explanation result. Such a method ignores the edges, which are critical for the cybersecurity analysts to investigate the alarm. The other three methods, i.e., GNNExplainer~\cite{GNNEx19}, PGExplainer~\cite{luo2020parameterized}, and GraphMask~\cite{schlichtkrull2021interpreting}, apply an \textit{edge-centric} strategy by identifying the important edges and regarding the constructed subgraph as the explanation result. Though the subgraph includes both important edges and nodes, the nodes identified in this way are usually not the truly important ones. Besides nodes and edges, only GNNExplainer investigates the important attributes. However, GNNExplainer identifies the important attributes globally, which is not specified for each node or edge.
}

\subsection{Requirement}

To accurately explain the GNN models, we believe an explanation method should satisfy the following requirements.

\begin{table}[t]
	\caption{Comparison of different GNN explanation methods (\CIRCLE = true; \Circle = false; \LEFTcircle = incomplete).}
	\centering
	\tabcolsep = .08cm
	\begin{tabular}{lcccc}
		\toprule
		{Method} & Node & Edge & Attribute & No Prior Knowledge \\ 
		\midrule
		{GNNExplainer} & \Circle & \CIRCLE & \LEFTcircle &  \CIRCLE \\
		{PGExplainer} & \Circle & \CIRCLE & \Circle & \Circle \\
		{GraphMask} & \Circle & \CIRCLE & \Circle & \Circle \\
		{PGM-Explainer} & \CIRCLE & \Circle & \Circle & \CIRCLE \\
		SubgraphX & \CIRCLE & \Circle & \Circle & \CIRCLE \\
		\textbf{{\name}} & \CIRCLE & \CIRCLE & \CIRCLE & \CIRCLE \\
		\bottomrule
	\end{tabular}
	\label{tb:explainers}
\end{table}

\textbf{Requirement \#1: comprehensive explanation.}
{\revision We derive the comprehensiveness from completeness in \cite{WarneckeAWR20}.} Particularly to GNNs, it refers to all the major factors in an input graph, which includes nodes, edges, and attributes. 
The factors in a cybersecurity-based graph are specially constructed from real situations. The information contained by different factors is learned and used by GNNs. 
The distrust in GNNs exists as long as the decision-making is not clear to the cybersecurity analysts. A comprehensive explanation for all the major factors is crucial for them to fully understand the GNN behaviors. 

\textbf{Requirement \#2: accurate explanation.}
An explanation is accurate if it is able to identify the important factors that contribute to the prediction. For an accurately identified subgraph, we assume that the prediction probability of it should be close to or even higher than its original prediction probability. 
If a prediction error is not precisely addressed, the same error may lead to vulnerability from malicious attacks. The inaccurate explanation would not be able to help diagnose the error but enlarge the vulnerability.


\textbf{Requirement \#3: no need for prior knowledge of GNN models.}
The cybersecurity models are not easily accessible due to two major reasons. {\revision First, the cybersecurity applications require more complex neural network architectures~\cite{WarneckeAWR20}.} Not only do the models consist of different types of neural networks, but the GNNs are adapted differently from basic GNNs. {\revision Second, in real scenarios, the users often times are using pre-trained models~\cite{hu2020pretraining} especially for complex models.} The prediction accuracy itself does not alleviate the distrust of a model from the users, due to the lack of transparency. 
Explanation methods without the need for prior knowledge are easier to access and utilize because of their flexibility. With the constraints, the explanation method with no prior knowledge requirement is preferred by cybersecurity analysts.

\begin{figure*}[t]\centering
	\includegraphics[width=0.9\textwidth]{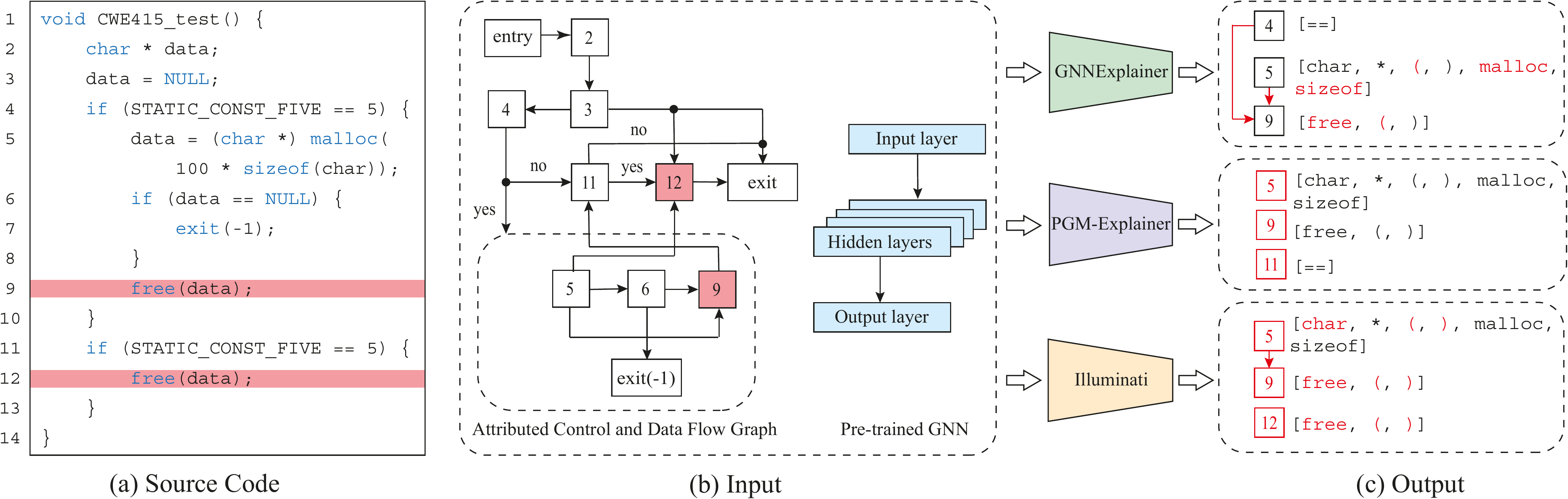}
	\caption{Explaining an example code predicted as vulnerable by a pre-trained GNN model with different explanation methods. (a) shows an example source code with ``double free'' vulnerability, (b) shows the converted \textbf{\underline{A}}ttributed control and data \textbf{\underline{F}}low \textbf{\underline{G}}raph (AFG) and a pre-trained model, and (c) shows the explanation results with the identified important factors colored. Specifically, GNNExplainer identifies important edges and treats the same attributes from different nodes identically, PGM-Explainer identifies important nodes only. 
	}
	\label{fig:comparison}
\end{figure*}

\subsection{Contribution}

Motivated by these, we design a comprehensive and accurate GNN explanation method, {\name}. 
Given a pre-trained GNN model and a graph as inputs, {\name} firstly learns the importance score of edges and node attributes collectively by using edge masks and attribute masks. {\name} then aggregates the learned masks and computes the importance score of nodes. In the end, our method identifies the important subgraph towards the GNN prediction. Attribute masks are applied locally to each attribute of each node so that we can identify the attributes that are important to different nodes. Further, {\name} does not require prior knowledge of the pre-trained model, which makes it more applicable to cybersecurity applications.

We compared the explanation performance of {\name} with prior works on public datasets and cybersecurity application datasets. We focused on two cybersecurity applications, i.e., smart contract vulnerability detection and code vulnerability detection. The evaluation is based on the prediction change between the input graph and the explained subgraph. 87.6\% {\name}-explained subgraphs retain their original prediction, with an improvement of 10.3\% over the baseline methods. 
Then we provided case studies for explaining the two real-world applications and a deep analysis of the model behaviors. 
We believe they can help cybersecurity analysts quickly understand and diagnose the alarms generated by applications using GNN models.

{\revision
In summary, we make three major contributions. 
\begin{itemize}
\item \textbf{New insight and method}. 
To the best of our knowledge, this is the first GNN explanation method to provide a comprehensive and cybersecurity-specialized explanation method for cybersecurity applications using GNN models. 

\item \textbf{Extensive evaluation.} 
We evaluate the performance of {\name} quantitatively with two cybersecurity applications. The results show {\name} outperforms existing explanation methods in terms of not only accuracy but also cybersecurity requirements.

\item \textbf{Cybersecurity case study.} 
We demonstrate the practical usage of {\name} with the case study of cybersecurity applications. We interpret the model behavior from both correct and incorrect predictions through the output of {\name}, as well as analyze how we can troubleshoot and improve the models.

\end{itemize}


The main novelty of {\name} is to jointly consider the contributions of nodes, edges, and attributes. Also, we analyze and prove that explaining node importance is critical for graph classification tasks.
Further, we find the node attributes should be explained individually for better comprehensiveness and accuracy. 

{\name} is different from existing works in terms of providing a comprehensive and accurate explanation method specialized for real cybersecurity applications. Particularly, compared with a representative related work, i.e., GNNExplainer, it is a generic method that only explains edges and does not explain node attributes individually.
}

{\revision
\section{Security Cases and Threat Models}
\label{sec_problem}


\subsection{Case \#1: Code Vulnerability}

\textbf{Code vulnerability} is the flaw or weakness in the code that can cause risks and be exploited by the attackers to conduct unauthorized activities, e.g., stealing data~\cite{chen2018internet}. For example, the straightforward risks of buffer overflow are data loss, software crashes, and arbitrary code execution, which can be exploited by attackers. A program is classified as vulnerable if it contains a vulnerability. The tested CWE dataset has three types of vulnerability: ``double free'', ``use after free'', and ``NULL pointer dereference''. 

\textbf{Threat model.}
The attackers can exploit the detected vulnerabilities to initialize malicious actions by using various attack patterns against the software or the system. The attackers can exploit the vulnerability simultaneously in different rewarding approaches, such as hacking tools and remote commands. These attacks may eventually lead to software crashes and data loss, profoundly, financial loss and privacy leakage. This impacts both users and developers.

\subsection{Case \#2: Smart Contract Vulnerability}
\textbf{Smart contract vulnerability} is a coding error that can be exploited by attackers to cause financial loss. A program with such a coding error is classified as vulnerable. Smart contract vulnerability is dangerous because most smart contracts deal with financial assets directly, and the blockchain cannot roll back changes. We study two types of vulnerabilities, i.e., reentrancy vulnerability and infinite loop vulnerability. The reentrancy vulnerability occurs when the contract transfers funds before the balance is updated. The infinite loop occurs when the loop never finishes.

\textbf{Threat model.}
The attackers can exploit the logical errors to conduct the attack by submitting a transaction to the blockchain. This can cause transaction failures or repeated transactions, which eventually lead to financial loss. For example, the malicious contract can drain funds from the reentrancy-vulnerable contract by recurrent re-entrant calls~\cite{Daniel_21}. The DAO attack is one exploitation case to such vulnerability. The attack conducts repeated withdrawals before the balance update. This attack has caused significant money stolen.
}

\section{Background}

\subsection{Graph Neural Networks}


\textbf{Graph Neural Networks (GNNs)}, $\Phi$ takes an attributed graph $G=(\mathcal{V}, \mathcal{E})$ and $\mathcal{X}$ as input then generates a set of node representations $\mathcal{Z}$ through hidden layers, where $\mathcal{V}$ and $\mathcal{E}$ denote nodes and edges, and $\mathcal{X}$ denotes attributes.

A GNN, $\Phi$ takes two major operations to compute node representations $\boldsymbol{h}$ in each layer~\cite{kipf2017semi, hamilton2017inductive, velickovic2018graph, ChenMX18}. In the $l$-th layer, GNN computes the neighbor representation $\boldsymbol{h}_{\mathcal{N}_{i}}^{(l)}=\textsc{Agg}(\{ \boldsymbol{h}_{j}^{(l-1)}\mid v_{j}\in \mathcal{N}_{i})\} )$ for node $v_{i}$ firstly, by aggregating its neighbor nodes' representations from the previous layer. Then, the new node representation is updated from the aggregated representation and its representation from the previous layer: $\boldsymbol{h}_{i}^{(l)}=\textsc{Update}(\boldsymbol{h}_{i}^{(l-1)}, \boldsymbol{h}_{\mathcal{N}_{i}}^{(l)})$. The final representation for node $v_{i}$ is $\boldsymbol{z}_{i}=\boldsymbol{h}_{i}^{(L)}$ after $L$ layers of computation. 
The final node representations are used for different tasks such as graph classification.
A generic graph classification model contains a pooling method and fully connected layers after GNN layers. The pooling method gathers node embeddings into a graph embedding and the fully connected layers compute the classification. 

In this paper, we design our explanation method based on the GNNs with such architecture, so our explanation method is more applicable. 

\subsection{GNN Explanation}


\textbf{GNN explanation} takes an attributed graph and a pre-trained GNN model as input, then identifies the key factors that contribute to the prediction. 
{\revision
Specifically, the task for the explanation methods is to identify the nodes, edges and attributes that contribute most to the prediction. }
For graph classification tasks, given an input graph $G$ with attributes $\mathcal{X}$ and a pre-trained GNN model $\Phi$, the GNN will make the prediction by computing the label $y$ with the probability $P_{\Phi}(Y=y\mid G, \mathcal{X})$. The task of explanation methods is to reason why the input graph is classified as $y$ by $\Phi$. The explanation offers a set of important factors that contribute to the prediction, for example, by retaining important edges~\cite{luo2020parameterized,schlichtkrull2021interpreting}.  

In this paper, we develop the explanation method {\name} for GNN models in cybersecurity domain. Existing works only focus on specific factors to explain. {\name} provides a comprehensive and accurate explanation for all the graph factors, which benefits the development of cybersecurity applications.

\textbf{Example with code vulnerability detection.}
Figure~\ref{fig:comparison}(a) shows an example source code with a ``double free'' vulnerability, which happens when the second \texttt{free} (line 12) is called after the first \texttt{free} (line 9). Vulnerability detection methods firstly convert the source code to an attributed graph. For example, we construct the attributed graph from the source code as shown in Figure~\ref{fig:comparison}(b) by building the {\revision \textbf{\underline{A}}ttributed control and data \textbf{\underline{F}}low \textbf{\underline{G}}raph (AFG)} and encoding the syntax attributes for each node. The node denotes the statement, the edge denotes control or data flow between two statements, and the attributes include syntax features, such as which keywords are used in a statement. Using the AFGs and their corresponding labels (benign or vulnerable) as the training dataset, one can train a GNN model for vulnerability detection, e.g.,  Devign~\cite{devign}. 

For the AFG generated from the example source code in Figure~\ref{fig:comparison}, nodes 9, 12 and the keyword \texttt{free} should be identified in the final explanation results. Figure~\ref{fig:comparison}(c) presents the output from two recent representative works and {\name}. 
GNNExplainer estimates the edge importance from the AFG by learning the soft continuous edge masks. In this example, GNNExplainer identifies $(4, 9)$ and $(5, 9)$ as important and considers this subgraph as the explanation result. This is not accurate because node 12 is missed due to none of its edges is considered important. 
PGM-Explainer samples a local dataset by random attribute perturbation to the AFG. With the perturbed nodes and the prediction change being recorded, a probabilistic graphical model is utilized to identify the important nodes. As a result, nodes 5, 9, and 11 are identified.
The explanation from PGM-Explainer misses node 12. 
Such explanations will confuse a cybersecurity analyst or lead to a wrong conclusion.

\begin{figure*}[t]
	\centering
	\includegraphics[width=0.85\textwidth]{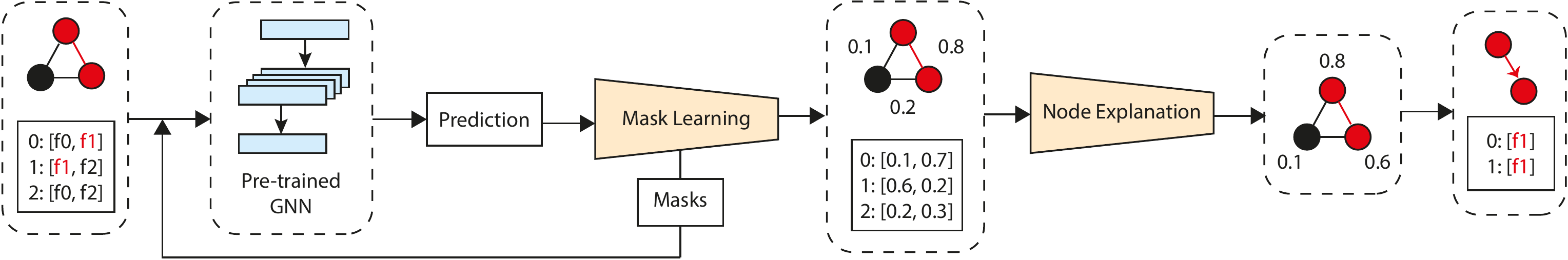}
	\caption{The workflow of {\name}. With a input graph and a pre-trained GNN, {\name} firstly learns the importance scores for edges and node attributes. Next, {\name} estimates the importance scores for nodes from the previous calculation. The important subgraph is then explained by removing the unimportant factors.}
	\label{fig:process}
\end{figure*}

\begin{table}[t]
	\caption{\revision List of notations.}
	\centering
	\tabcolsep = .5cm
    {\revision
	\begin{tabular}{c|l}
		\toprule
		Notation & Description \\ 
		\midrule
		$G$ & A graph \\
		$\mathcal{V}$ & The set of nodes in graph $G$ \\
		$\mathcal{E}$ & The set of edges in graph $G$ \\
		$\mathcal{X}$ & The sets of node attributes in graph $G$ \\
		$\Phi$ & A GNN model \\
		$P$ & Prediction probability \\
		$m$ & Explanation mask \\ 
		$\omega$ & Importance score \\
		\bottomrule
	\end{tabular}
	\label{tb:notations}
    }
\end{table}%

\section{Design Details of {\name} }
\label{sec_gnn_explanation}



\subsection{Overview}



The workflow of {\name} is shown in Figure~\ref{fig:process}. {\name} takes an attributed graph and a pre-trained GNN as input then generates a key subgraph that contributes to the prediction, with the importance scores as the importance measurement. 

First, {\name} learns the importance scores for edges and node attributes collectively from the input graph and the pre-trained GNN. The edge masks and attribute masks are initialized by {\name}. {\revision Using the same approach from GNNExplainer, {\name} applies the masks as learnable parameters to the input graph. Similar to GNN training, the masks are learned iteratively from the feedback of GNN.} The importance scores are then calculated from the learned masks.
Next, {\name} estimates the importance scores for nodes from the calculated importance scores for edges and node attributes. For each node, the importance scores from the related edges and attributes are aggregated for the estimation. 
Finally, an important subgraph is explained by removing the factors with low importance scores under certain constraints, e.g., the size of the subgraph.

Next, we discuss the detailed design of {\name}. The main notations are summarized in Table~\ref{tb:notations}


\subsection{Objective Function}



An attributed graph contains graph structure and attributes. Our target is to find a subgraph $G_{s}=(\mathcal{V}_{s}, \mathcal{E}_{s})$ and a subset of attributes $\mathcal{X}_{s}$ that contribute to the GNN prediction. In order to find the important factors, we use mutual information maximization as our objective function~\cite{GNNEx19}, which is defined in Equation~\ref{eq:mi}:
\begin{align}
\label{eq:mi}
\begin{aligned}
	& \max_{G_{s}} MI(Y, (G_{s}, \mathcal{X}_{s}))= H(Y) - \\
	& \qquad H(Y\mid G=G_{s}, \mathcal{X}=\mathcal{X}_{s})
\end{aligned}
\end{align}
where $Y$ is the predicted label for an input graph. The graph structure can be represented by an adjacency matrix $A$ or an edge list $\mathcal{E}$, and node attributes are represented by a node attribute matrix. However, a node consists of its connected edges and attributes.
It is not possible to directly quantify the importance score for a node. Thus, node explanation is considered after edge and node attribute explanation. Here, $G_{s} = (\mathcal{V}, \mathcal{E}_{s})$. 

\textbf{Estimation for edges.} The estimation for the objective function is not tractable since there are $2^{\mid \mathcal{E}\mid}$ different subgraphs for $G$, because each edge is independent. Following the existing works~\cite{GNNEx19, luo2020parameterized}, in consideration of relaxation, we adopt Bernoulli distribution $P(G_{s})=\prod_{(i,j)\in \mathcal{E}}P((i,j))$ for edge explanation, where $P((i,j))$ is the probability of the edge $(i,j)$'s existence. Therefore, our goal for edge explanation is considered as finding the correct $P(G_{s})$.

\textbf{Estimation for attributes.} For the basic GNNs, the same node attributes from different nodes share the same GNN parameters in each layer, while some newly developed GNNs extend the usage of node attributes. For example, GAT~\cite{velickovic2018graph} takes node attributes to calculate attention coefficients. Besides, the same node attributes perform differently when located in different nodes because of the nonlinear computation from GNNs. Node attributes should be explained individually for a graph. We use the same method from edge estimation for node attribute estimation. 

The mutual information quantifies the probability change of GNN prediction with the input limited to $G_{s}$ and $\mathcal{X}_{s}$. An edge $(i,j)$ is considered unimportant when removing it does not largely decrease the probability of prediction. With the pre-trained GNN $\Phi$ being fixed, we rewrite our objective function as minimizing $H(Y\mid G=G_{s}, \mathcal{X}=\mathcal{X}_{s})$, defined in Equation~\ref{eq:obj}, where $C$ is the set of prediction classes. In this way, we make sure the subgraph $G_{s}G_{s}$ and the subset of attributes $\mathcal{X}_{s}$ achieve the maximum probability of prediction.
\begin{align}
\label{eq:obj}
\begin{aligned}
	& \min_{P(G_{s}),P(\mathcal{X}_{s})}-\sum_{c=1}^{C}\mathbbm{1}[y=c]
	\log P_{\Phi}(Y=y\mid \\
	& \qquad G=G_{s}, \mathcal{X}=\mathcal{X}_{s})
\end{aligned}
\end{align}


\subsection{Edge and Attribute Explanation}

%
%

Our goal for edge and attribute explanation is to learn the correct $P(G_{s})$ and $P(\mathcal{X}_{s})$. We introduce edge masks $m^{(\mathcal{E})}$ and node attribute masks $m^{(\mathcal{X})}$ as our learning parameters. We take $P(G_{s})=\sigma (m^{(\mathcal{E})})$ and $P(\mathcal{X}_{s})=\sigma (m^{(\mathcal{X})})$, where $\sigma (\cdot)$ denotes \textit{sigmoid} function. Here, the objective function can be approximated as:
\begin{align}
\label{eq:obj_mask}
\begin{aligned}
	& \min_{m^{(\mathcal{E})},m^{(\mathcal{X})}}
	-\sum_{c=1}^{C}\mathbbm{1}[y=c]	\log P_{\Phi}(Y=y\mid G= \\
	& \qquad (\mathcal{V}, \mathcal{E}\odot \sigma (m^{(\mathcal{E})})), 
	\mathcal{X}=\mathcal{X}_{s}\odot \sigma (m^{(\mathcal{X})}))
\end{aligned}
\end{align}
where $\odot$ denotes element-wise multiplication. Edge masks learn how much message from source nodes should be passed to destination nodes. Node attribute masks learn how much of node attributes should be used for messages. 

For undirected graphs, the edge is bidirectional, where the information is passed back and forth. In this paper, we consider all the graphs as directed graphs to estimate the message passing precisely. An edge mask for undirected graph is computed by $\hat{m}_{(i,j)}^{(\mathcal{E})}=\hat{m}_{(j,i)}^{(\mathcal{E})}=Agg(\{ \hat{m}_{(i,j)}^{(\mathcal{E})}, \hat{m}_{(j,i)}^{(\mathcal{E})}\} )$, where $Agg$ is a user-defined aggregation function. GNNExplainer and PGExplainer treat both directions equally by taking the average of two directions. From our practical observation, the performance of the explanation can be improved by applying different aggregation functions. 

As Figure~\ref{fig:process} suggests, mask training is similar to GNN training. 
First, we initialize the masks for edges and node attributes, respectively. Next, the masks are used to add weights on the edges and node attributes of the input graph as in Equation~\ref{eq:obj_mask}. Then, the weighted graph is fed into the pre-trained GNN for mask learning. With the feedback from GNN, the mask values are optimized by minimizing the objective function. The masks are learned iteratively through these steps, so the importance scores are gathered from the learned masks. 

\textbf{Reparameterization trick.}
The importance scores, as weights for mask training, are soft continuous values falling into $(0,1)$. However, an edge should either exist or not, meaning the edges should be binarily indicated. 
Using continuous importance scores will cause the ``introduced evidence'' problem~\cite{yuan2021explainability}. The importance scores add unexpected noise to the input, which does not reflect the real-world explanations. The binary importance scores, however, are not differentiable for researchers to estimate the importance level.
Our solution is to reparameterize importance scores into binary as weights on the input graph, while the differentiable importance scores are still retained for importance estimation. Here, we apply hard concrete distribution~\cite{louizos2018learning} as our reparameterization trick. We rewrite the distribution for edges as:
\begin{equation}
\label{eq:hard_mask}
\begin{aligned}
s &= \sigma ((\log u - \log (1-u) + m^{(\mathcal{E})}) / \beta )\\
\epsilon &= \min (1, \max (0, s(\zeta - \gamma )+\gamma ))
\end{aligned}
\end{equation}
where $u\sim \mathcal{U}(0,1)$ and $\beta$ is the temperature. With $\zeta < 0$ and $\gamma > 1$, we stretch the concrete distribution to $(\zeta , \gamma )$. Distribution in $(\zeta , 0]$ and $[1, \gamma )$ ultimately falls into 0 and 1. Thus, part of the distribution is squeezed into binary. Meanwhile, we take $s = \sigma (m^{(\mathcal{E})} / \beta )$ as the binary concrete distribution for edges, i.e., importance scores, then approximate the ``sub-edges'' as $\mathcal{E}_{s} \approx \mathcal{E}\odot \epsilon$ for edge mask training.

\subsection{Node Explanation}

With learned edge masks and node attribute masks, we need to quantify the importance scores for nodes. Inspired by the Bernoulli distribution for graph structure, the contribution from a node $v_{i}$ is quantified by:
\begin{equation}
\label{eq:prob_node}
	\omega _{v_{i}}=
	\prod_{(i,j)\in \mathcal{E}_{i}^{+}} P((i,j))^{1/\mid \mathcal{E}_{i}^{+}\mid }
	\prod_{t\in \boldsymbol{x}_{i}} P(t)^{1/\mid x_{i}\mid}
\end{equation}
Here, the contribution of a node $v_{i}$ is quantified from the importance scores of its outgoing edges $\mathcal{E}_{i}^{+}$ and node attributes $\boldsymbol{x}_{i}$.
The contribution from edges should be normalized because a node connects arbitrary numbers of edges. We multiple the importance scores of connected edges and extract the $\mid \mathcal{E}_{i}^{+}\mid$-th root of the multiplication.
For node $v_{i}$, we can define the importance score for outgoing edges as $\omega _{\mathcal{E}_{i}^{+}}=\prod_{(i,j)\in \mathcal{E}_{i}^{+}} P((i,j))^{1/\mid \mathcal{E}_{i}^{+}\mid }$, and the importance score for node attributes as $\omega _{\boldsymbol{x}_i}=\prod_{t\in \boldsymbol{x}_{i}} P(t)^{1/\mid x_{i}\mid}$.
However, there are two problems with equation~\ref{eq:prob_node}. First, the normalization method may degrade the important edges. An important node can be connected by important and unimportant edges while the unimportant edges decrease the overall importance of its message passing path. Second, node interactions are not considered. Nodes interact through GNN computation, which leads to certain nodes being important to the prediction. 

In order to fix the first problem, we take an aggregation function, e.g., $\max$, to calculate the contribution from $\mathcal{E}_{i}^{+}$, $\omega _{\mathcal{E}_{i}^{+}}=Agg(\{ P((i,j))\mid (i,j) \in \mathcal{E}_{i}^{+}\})$. The aggregation function is changeable in order to adjust to different GNNs. But it cannot be directly applied to the incoming edges of $v_{i}$. In GNN computation, a node's representation $\boldsymbol{h}_{i}$ is aggregated from the message passing through its incoming edges $\mathcal{E}_{i}^{-}$. The message information depends on the source node and its connected edge. Thus, we quantify the message importance through edge $(i,j)$ as:
\begin{equation}
\label{eq:msg}
	\omega _{(i,j)} = P((i,j)) \omega _{\boldsymbol{x}_i}
\end{equation}
For a node's importance estimation, we consider the messages from and to the node (outgoing messages and incoming messages) separately since the contribution can vary. With the solution to the first problem, we firstly aggregate the importance scores for outgoing messages and incoming messages of node $v_{i}$ separately:
\begin{equation}
\label{eq:msg_2}
\begin{aligned}
	\omega _{v_{i}}^{(out)} &= Agg_{1}
	(\{ \omega _{(i,j)}\mid (i,j) \in \mathcal{E}_{i}^{+}\} )\\
	\omega _{v_{i}}^{(in)} &= Agg_{1}
	(\{ \omega _{(j,i)}\mid (j,i) \in \mathcal{E}_{i}^{-}\} ) 
\end{aligned}
\end{equation}
Then, we introduce the second aggregation function to compute the ultimate node importance scores from gathering the outgoing messages and incoming messages. Here, we compute the ultimate importance score for $v_{i}$ by:
\begin{equation}
\label{eq:w_node}
	\omega _{v_{i}} = Agg_{2}(\{ \omega _{v_{i}}^{(out)}, \omega _{v_{i}}^{(in)}\} )
\end{equation}

\textbf{Synchronized mask learning. }For different purposes, some graph factors can share the same masks. For example, for undirected graphs, two paths of the same edge can share the same edge mask in order to eliminate the pair difference problem. 
When node attribute explanation is not required, node attributes from the same node $\boldsymbol{x}_i$ can share the same masks. In this way, we are able to directly learn $\omega _{\boldsymbol{x}_i}$ for each node. Thus, the graph structure is explained efficiently with less storage requirement.

\begin{table}[t]
	\caption{The specifications of different dataset and the accuracy of the pre-trained models.}
	\label{tb:setup}
	\centering
	\tabcolsep = 0.18cm
	\begin{tabular}{lcccc}
		\toprule
		\textbf{Dataset} & \makecell{\textbf{Avg. \# }\\\textbf{of nodes}} & \makecell{\textbf{\# of train/}\\\textbf{validation/test}} & \textbf{Model} & \textbf{Accuracy} \\ 
		\midrule
		BBBP & 24.065 & 1,629/205/205 & GCN & 0.878\\
		Mutagenicity & 30.317 & 3,467/435/435 & GCN & 0.805\\
		BA-2motifs & 25.000 & 800/100/100 & GCN & 1.000\\
		Reentrancy & 4.968 & 1,340/$\cdot$/331 & DR-GCN & 0.926\\
		Infinite Loop & 3.686 & 1,056/$\cdot$/261 & DR-GCN & 0.632\\
		CWE-415 & 9.962 & 666/$\cdot$/334 & Devign & 0.949\\
		CWE-416 & 17.839 & 666/$\cdot$/334 & Devign & 0.934\\
		CWE-476 & 9.132 & 666/$\cdot$/334 & Devign & 0.841\\
		\bottomrule
	\end{tabular}
\end{table}%


\section{Experiment}
\label{sec_experiment}


The experiments are conducted on a server with two Intel Xeon E5-2683 v3 (2.00GHz) CPUs, each of which has 14 cores and 28 threads. The code in this work is available for reproduction\footnote{\url{https://github.com/iHeartGraph/Illuminati}}.

\subsection{Dataset and Pre-traind GNN Models}

\begin{table}[t]
\caption{EP (\%) of explained subgraphs for public datasets, where BBBP and Mutagenicity are real-world molecular datasets and BA-2motifs is a synthetic dataset.}
\label{tb:general_result}
\centering
\tabcolsep = 0.3cm
\begin{tabular}{lrrr}
\toprule
\textbf{Methods} & \textbf{BBBP} & \textbf{Mutagenicity} & \textbf{BA-2motifs}\\ 
\midrule
PGM-Explainer & 74.6 & 57.2 & 41.0 \\
GNNExplainer & 75.1 & 69.9 & 41.0 \\ 
PGExplainer & 76.2 & 68.2 & 41.0 \\
\textbf{{\name}} & \textbf{76.7} & \textbf{72.0} & 41.0 \\
\bottomrule
\end{tabular}
\end{table}



We evaluate eight datasets as shown in Table~\ref{tb:setup}. 
We test the explanation methods on three public datasets used for the graph classification task, including two real-world datasets and a synthetic dataset. Two molecular datasets Mutagenicity~\cite{mutag} and BBBP~\cite{Ramsundar-et-al-2019} contain graphs with nodes representing the atoms, and edges representing the chemical bonds. BA-2motifs~\cite{luo2020parameterized} is a motif-based synthetic dataset, each graph from which contains a five-node house-like motif or a cycle motif.
For code vulnerability detection, we use a well-labeled dataset from NIST Software Assurance Reference Dataset (SARD), named Juliet~\cite{nistsard}, which not only labels the vulnerable functions but also provides the benign functions. For a clear explanation study, we require the datasets easy to understand and achieve good prediction accuracy. The CWEs we select for the experiment are 415, 416, and 476 which represent ``double free'', ``use after free'' and ``NULL pointer dereference'' respectively. The source code is represented by AFGs.
The datasets for smart contract vulnerability detection are from two platforms, Ethereum Smart Contracts (Reentrancy) and VNT chain Smart Contracts (Infinite loop). The contract graphs are constructed from the source code from the work of Zhuang \textit{et al.}~\cite{ijcai2020-454}.
The graphs for two cybersecurity applications will be illustrated in Section~\ref{sec_case_study}.


We use three kinds of GNN models for different applications respectively. The models include two parts, i.e., GNN layers to generate node representations and functional layers to compute graph representations.
The dataset splits for model training and the testing accuracies are shown in Table~\ref{tb:setup}. The pre-trained models are used as pre-trained models for explanation evaluation. 

We train a basic 3-layer GCN~\cite{kipf2017semi} for public datasets. For a graph classification task, it is followed by a $max$ and $mean$ pooling layer and a fully connected layer. 
The model in Devign~\cite{devign} is used for code vulnerability detection, which consists of a 3-layer gated graph recurrent network~\cite{ggnn} with a \textit{Conv} module.
DR-GCN~\cite{ijcai2020-454} for smart contract vulnerability detection is derived from GCN with increased connectivity in each layer. A \textit{max} pooling layer and two fully connected layers are applied for graph representation after the 3-layer DR-GCN.

\subsection{Compared Works}

We compare {\name} with the following baseline GNN explanation methods, GNNExplainer~\cite{GNNEx19}, PGM-Explainer~\cite{pgmexplainer}, and PGExplainer~\cite{luo2020parameterized}. Here, GNNExplainer and PGM-Explainer do not require prior knowledge from GNNs.
GNNExplainer targets on edges for graph structure explanation. The importance of edges is differentiated by learning the edge masks. The important nodes are automatically extracted with the explained important edges. Attribute explanation is also provided by GNNExplainer. The same node attributes from different nodes are explained equally by learning the same attribute masks. 
PGM-Explainer~\cite{pgmexplainer} provides node explanation by a probabilistic graphical model with the generated dataset. Whether a node is perturbed and the prediction change is noted for dataset generation. Then the Grow-Shrink (GS)~\cite{NIPS1999_GS} algorithm is conducted to shrink the datasets and a Bayesian network is used to explain the GNN model.
PGExplainer takes the node embeddings from the last layer of GNNs as input, then learns the edge masks from a multi-layer neural network. Similar to GNNExplainer, the explanation of graph structure is only determined by explained edges.

We used the shared source code of the two compared works and reimplement the interfaces to support the dataset and pre-trained GNN models.
We compare different methods for graph structure explanation. 
Specifically, the subgraph is extracted only by node, and all the connected edges are retained. For GNNExplainer and PGExplainer, as we identify the top-$R$ (rate) or top-$K$ nodes, edges that are originally connected from the input graph are restored. Thus, only node removal is conducted and the number of remaining nodes is controlled to be equal for all the explanation methods. Also, we do not apply any additional constraints for the evaluation. We use $max$ pooling as $Agg^{(2)}$ for {\name}.

\begin{table*}[t]
\caption{EP (\%) of explained subgraphs. $R=0.5$ for smart contract vulnerability detection; $K=6$ for code vulnerability detection.}
\label{tb:ep+}
\centering
\tabcolsep = 0.5cm
\begin{tabular}{lrrrrr}
\toprule
\textbf{Methods} & \textbf{Reentrancy} & \textbf{Infinite loop} & \textbf{CWE-415} & \textbf{CWE-416} & \textbf{CWE-476} \\ 
\midrule
PGM-Explainer &
61.3 & 58.6 & 79.6 & 74.3 & 72.2 \\
GNNExplainer & 
81.3 & 72.0 & 81.7 & 74.9 & 85.0 \\
PGExplainer &
84.9 & 73.6 & \textbf{90.1} & 77.2 & 92.2 \\
\textbf{{\name}} & 
\textbf{93.4} & \textbf{78.2} & 88.0 & \textbf{80.8} & \textbf{97.3} \\
\bottomrule
\end{tabular}
\end{table*}

\begin{figure*}[!t]
\centering
\begin{subfigure}{.5\linewidth}
\centering
\includegraphics[width=\linewidth]{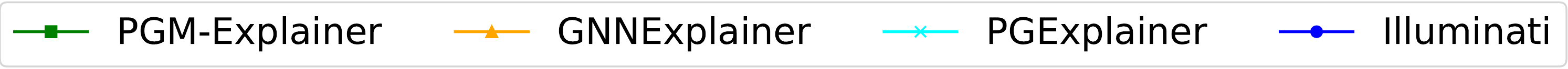}%
\end{subfigure}\\%
\begin{subfigure}{.2\linewidth}
\centering
\includegraphics[width=\linewidth]{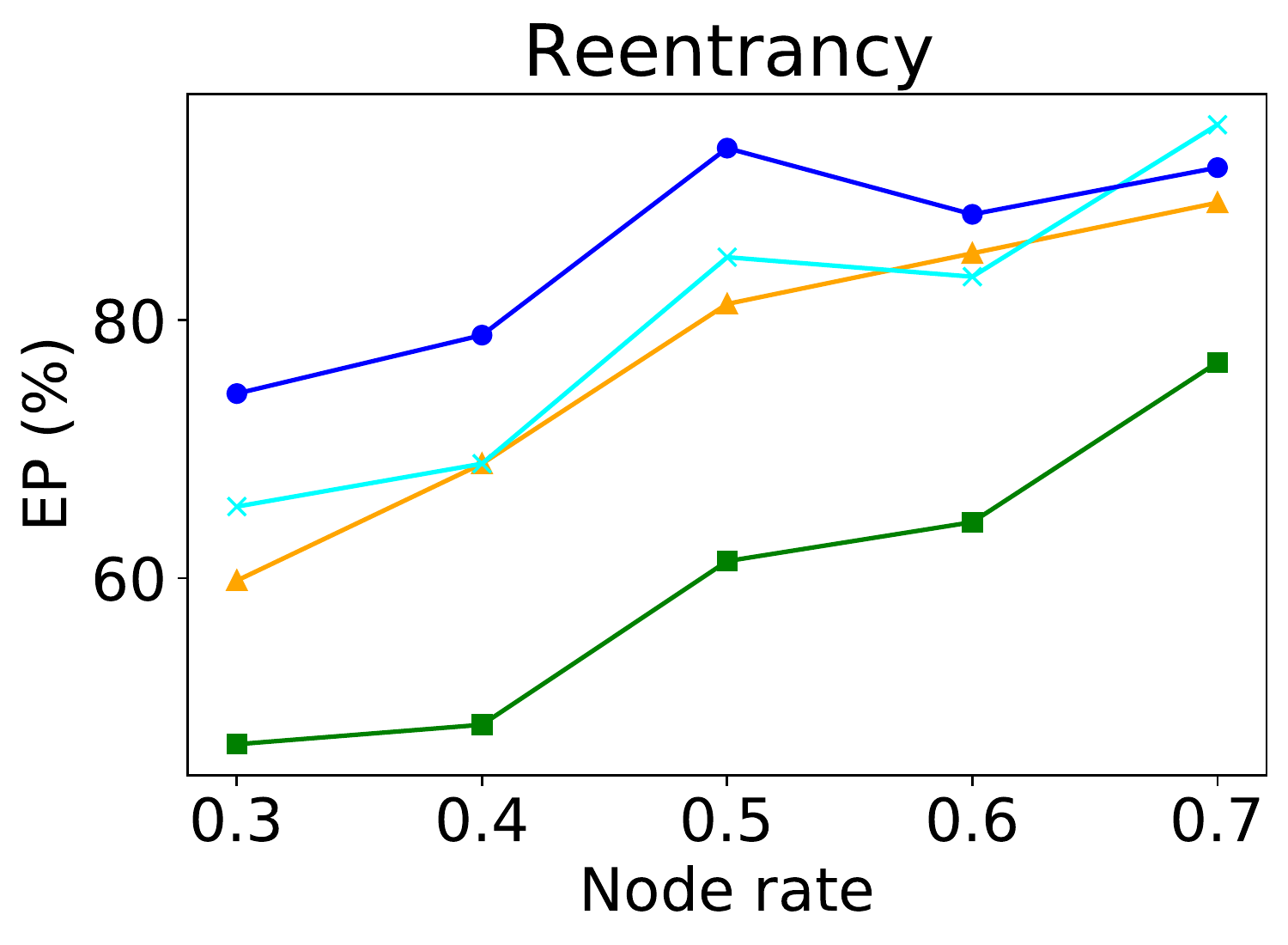}
\end{subfigure}%
\begin{subfigure}{.2\linewidth}
\centering
\includegraphics[width=\linewidth]{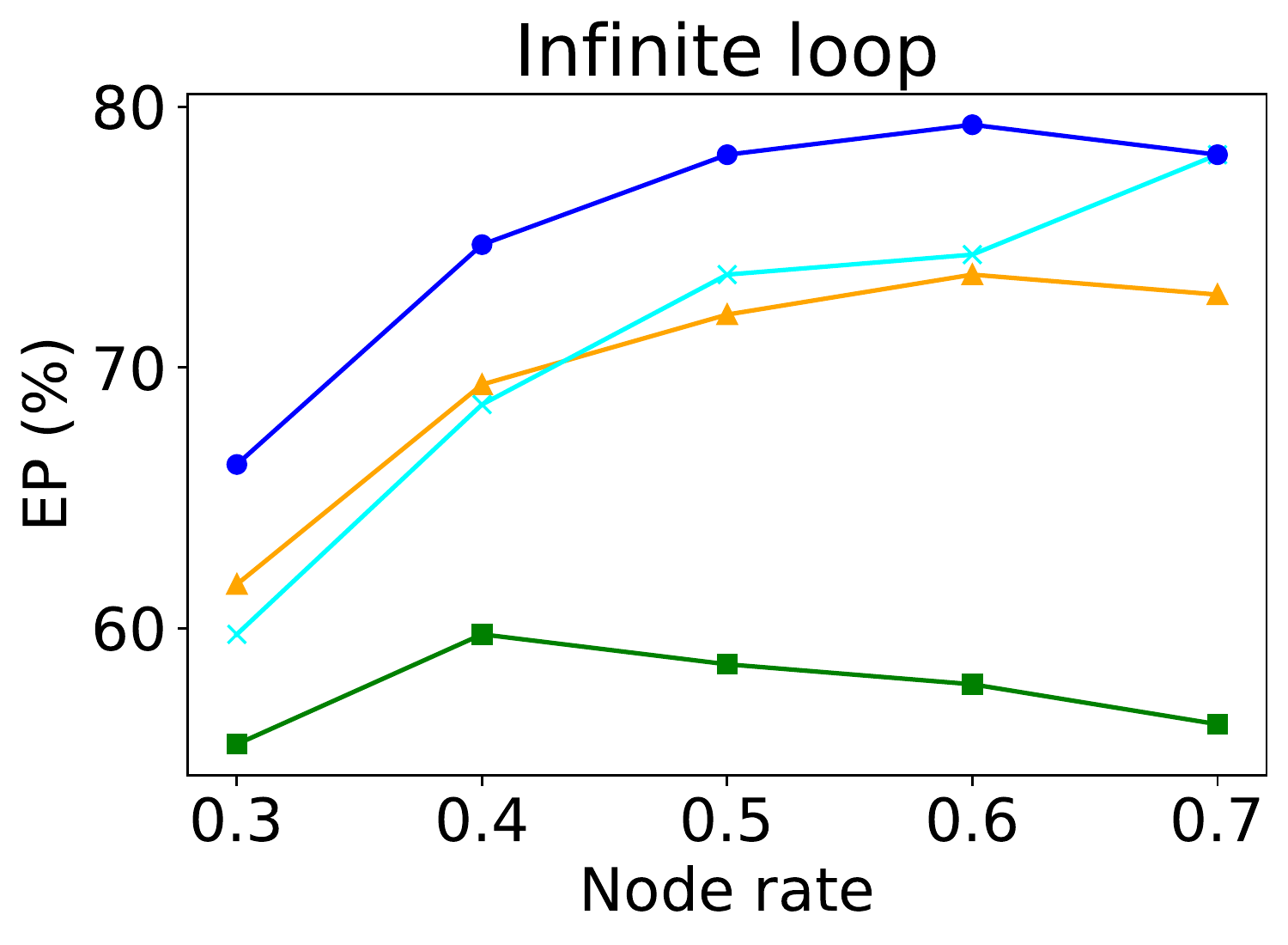}%
\end{subfigure}%
\begin{subfigure}{.2\linewidth}
\centering
\includegraphics[width=\linewidth]{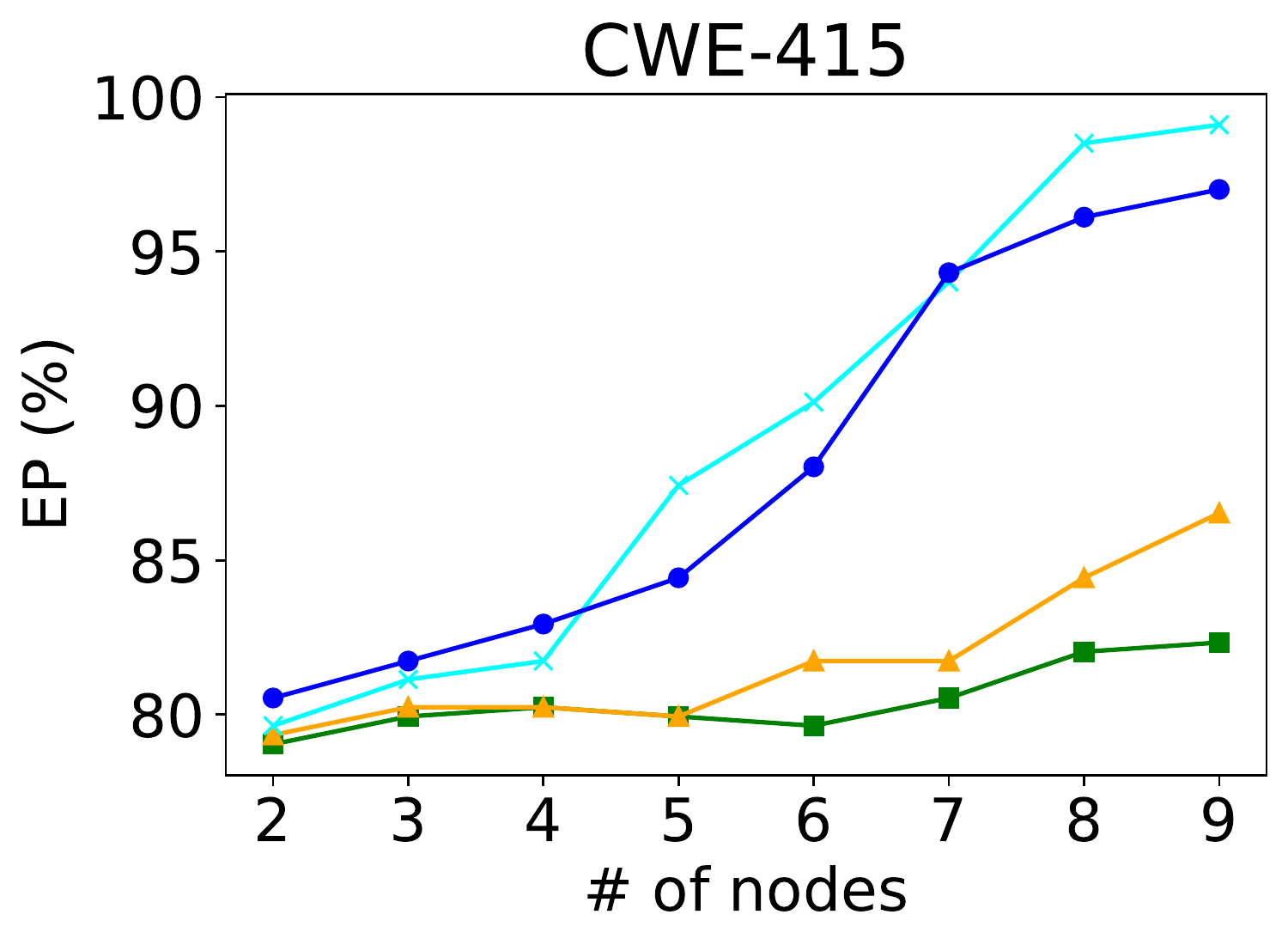}
\end{subfigure}%
\begin{subfigure}{.2\linewidth}
\centering
\includegraphics[width=\linewidth]{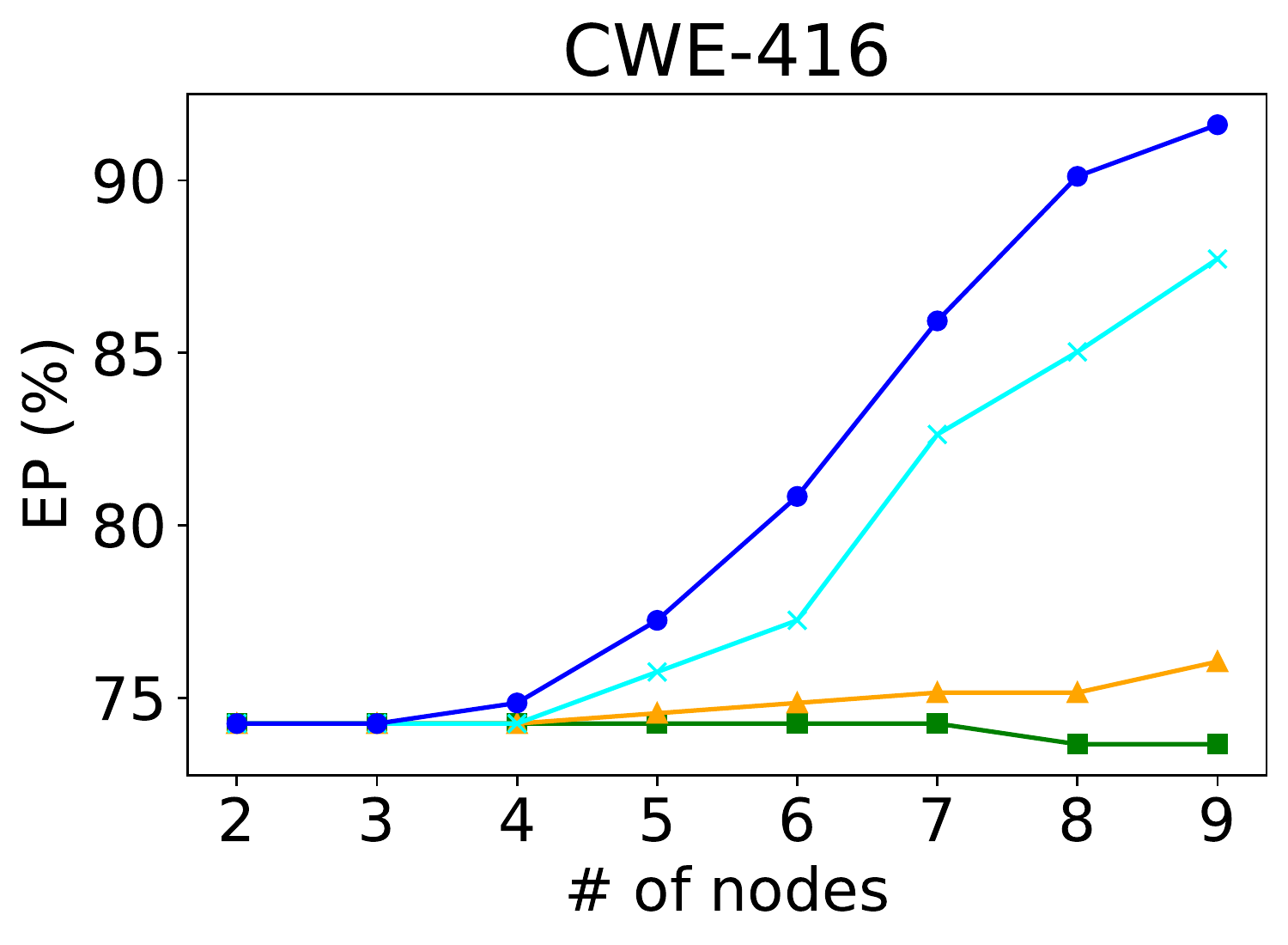}
\end{subfigure}%
\begin{subfigure}{.2\linewidth}
\centering
\includegraphics[width=\linewidth]{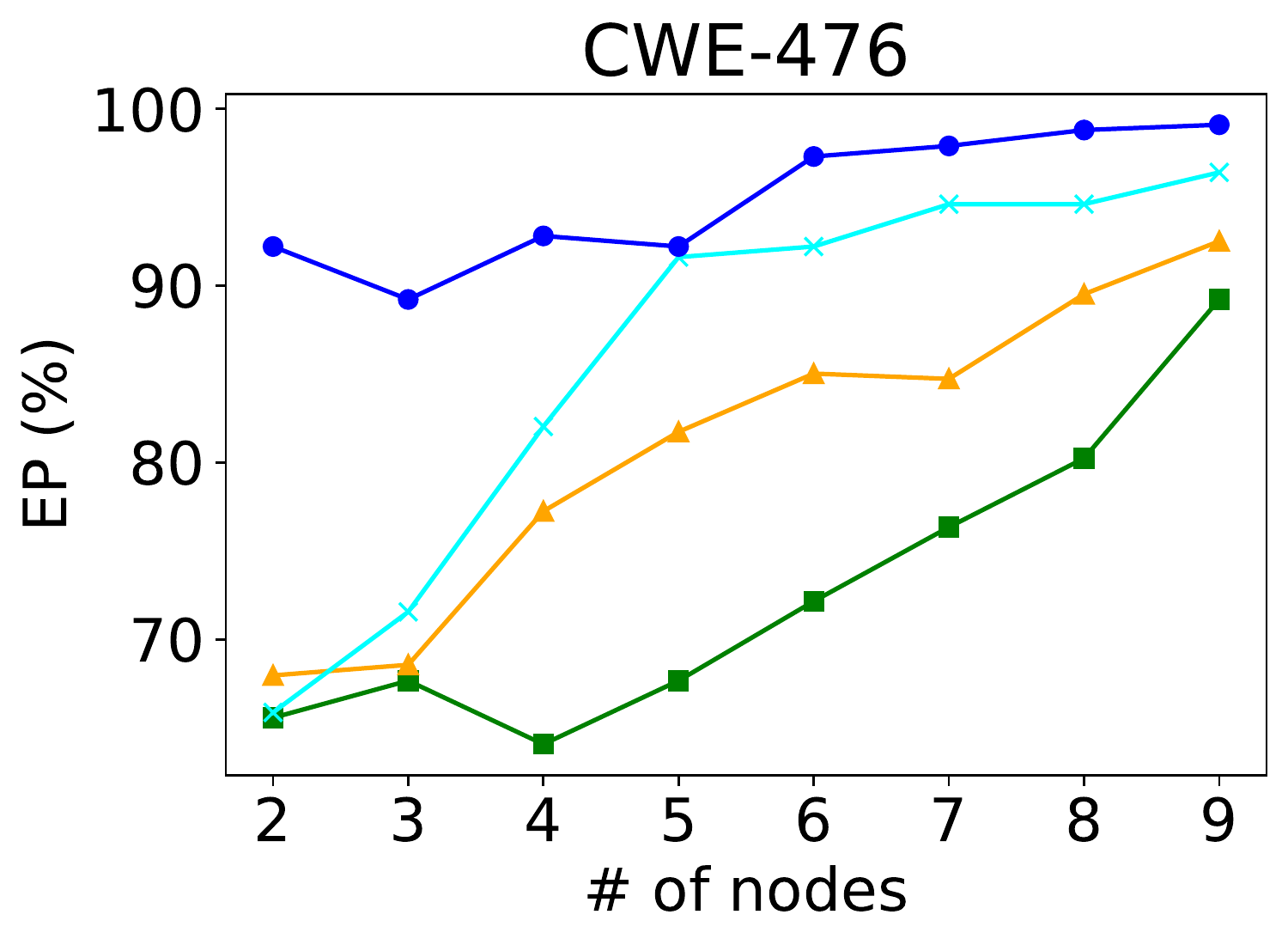}
\end{subfigure}%
\caption{Explanation results for cybersecurity applications. We obtain EP of explained subgraphs by changing explained subgraph size.}
\label{fig:ep+}
\end{figure*}

\subsection{Performance Comparison}


In this subsection, we present the quantitative analysis of explanation methods with various evaluation metrics.

\textbf{Evaluation metrics.}
In this work, we assume the important subgraphs will retain the original predictions, meaning causing the least prediction change from the original graphs. We define \textit{Essentialness Percentage (EP)} as our evaluation metric:
\begin{equation}
	\text{EP} = \frac{1}{N}\sum_{i}^{N}(\mathbbm{1}[y_{s}^{(i)}=y^{(i)}])
\end{equation}
where $\mathbbm{1}[\cdot]$ means the result being 1 if the statement in $[\cdot]$ is true, otherwise 0; $y_{s}$ denotes the prediction label of the subgraph, and $N$ is the number of graphs in the dataset. EP, as the percentage of subgraphs that retain the original predictions, evaluates how accurate the extracted factors are to the prediction. 
To validate the accuracy of the explained factors, we design two tests. Based on the objective of explanation, we firstly evaluate EP from the subgraphs formed by the important factors. We also consider the intuition reasonable that if the important factors are removed, the remaining subgraphs will not likely be able to retain the original predictions, which will cause lower EP. Thus, we divide the graphs into the explained subgraphs and the remaining subgraphs after explanation, where the explained subgraphs are constituted by important factors. 

An accurate explanation should be able to identify the most important factors, thus the explanation should be sparse. However, explanation methods provide continuous importance scores for different factors rather than solid binary scores. In order to evaluate the sparsity for different explanation methods, we define $Sparsity$ as follows:
\begin{equation}
	Sparsity = \frac{1}{N}\sum_{i}^{N} \text{min} \mid \mathcal{V}_{s}^{(i)} \mid
	\; \text{s.t.}\; y_{s}^{(i)}=y^{(i)}
\end{equation}
Sparsity represents the average minimum size of subgraphs that retain the original GNN predictions from a dataset. The smaller sparsity means the explanation method identifies more important factors and ignores irrelevant factors, thus provides more accurate explanations.

\textbf{EP of explained subgraphs.}
We use the testing splits from Table~\ref{tb:general_result} for explanation method evaluation. All the explanation methods explain the graph by generating the importance scores for different factors. It is unclear if a factor should be kept. Thus, we evaluate the performance of the explanation by comparing the EP under the same graph size.
First, we test the explanation methods with public datasets and a trained basic 3-layer GCN, shown in Table~\ref{tb:general_result}. We extract the top-10 nodes for Mutagenicity and BBBP, and the top-5 for synthetic dataset BA-2motifs. The result suggests that PGExplainer, as an explanation method requiring prior knowledge, outperforms other compared methods without prior knowledge. Overall, the explanation result shows that {\name} achieves the best EP in real-world datasets and outperforms other explanation methods. 

The explanation results for two cybersecurity applications are shown in Figure~\ref{fig:ep+}. Table~\ref{tb:ep+} summarizes the result values in the middle from Figure~\ref{fig:ep+}. As for smart contract detection, we variate the rate of extracted nodes; and we change the number of extracted nodes in code vulnerability detection. If the graph size to be explained is larger than the input graph size, then this graph is not considered for evaluation.

In general, {\name} shows the highest EP among other explanation methods in both applications, meaning it identifies the important subgraphs more accurately. For real-world datasets, PGM-Explainer does not perform as well as public datasets and synthetic datasets. The real-world datasets contain a more arbitrary and larger size of node attributes. PGExplainer outperforms other explanation methods in CWE-415, while the performance of {\name} is close to PGExplainer. 
To acquire better explanation accuracy, PGM-Explainer should be executed as the size of subgraphs changes; while GNNExplainer and {\name} only need to be executed once. As an explanation method that requires prior knowledge of GNNs, the performance of PGExplainer is generally better than the peer explanation methods without prior knowledge. However, without exploring nodes in depth, PGExplainer generally does not gain a higher EP than {\name}.
The result also suggests that as the size of explained subgraphs increases, the explanation is more accurate. We use real-world datasets, which ensure a node should not have an extremely high or low contribution. The predictions rely on the interactions between different nodes. 

\begin{table*}[t]
\caption{EP (\%) of remaining subgraphs for cybersecurity applications. $R=0.5$ for smart contract vulnerability detection; $K=6$ for code vulnerability detection.}
\label{tb:ep-}
\centering
\tabcolsep = 0.5cm
\begin{tabular}{lrrrrr}
\toprule
\textbf{Methods} & \textbf{Reentrancy} & \textbf{Infinite loop} & \textbf{CWE-415} & \textbf{CWE-416} & \textbf{CWE-476} \\ 
\midrule
PGM-Explainer &
76.1 & 70.1 & 86.5 & 85.6 & 85.3 \\
GNNExplainer & 
63.1 & 56.7 & 83.2 & 79.6 & 72.8 \\
PGExplainer &
72.2 & 59.8 & \textbf{72.2} & \textbf{59.9} & 59.6 \\
\textbf{{\name}} & 
\textbf{51.7} & \textbf{58.2} & \textbf{72.2} & 62.0 & \textbf{49.4} \\
\bottomrule
\end{tabular}
\end{table*}

\begin{figure*}[!t]
\centering
\begin{subfigure}{.5\linewidth}
\centering
\includegraphics[width=\linewidth]{./picture/legend.pdf}%
\end{subfigure}\\%
\begin{subfigure}{.2\linewidth}
\centering
\includegraphics[width=\linewidth]{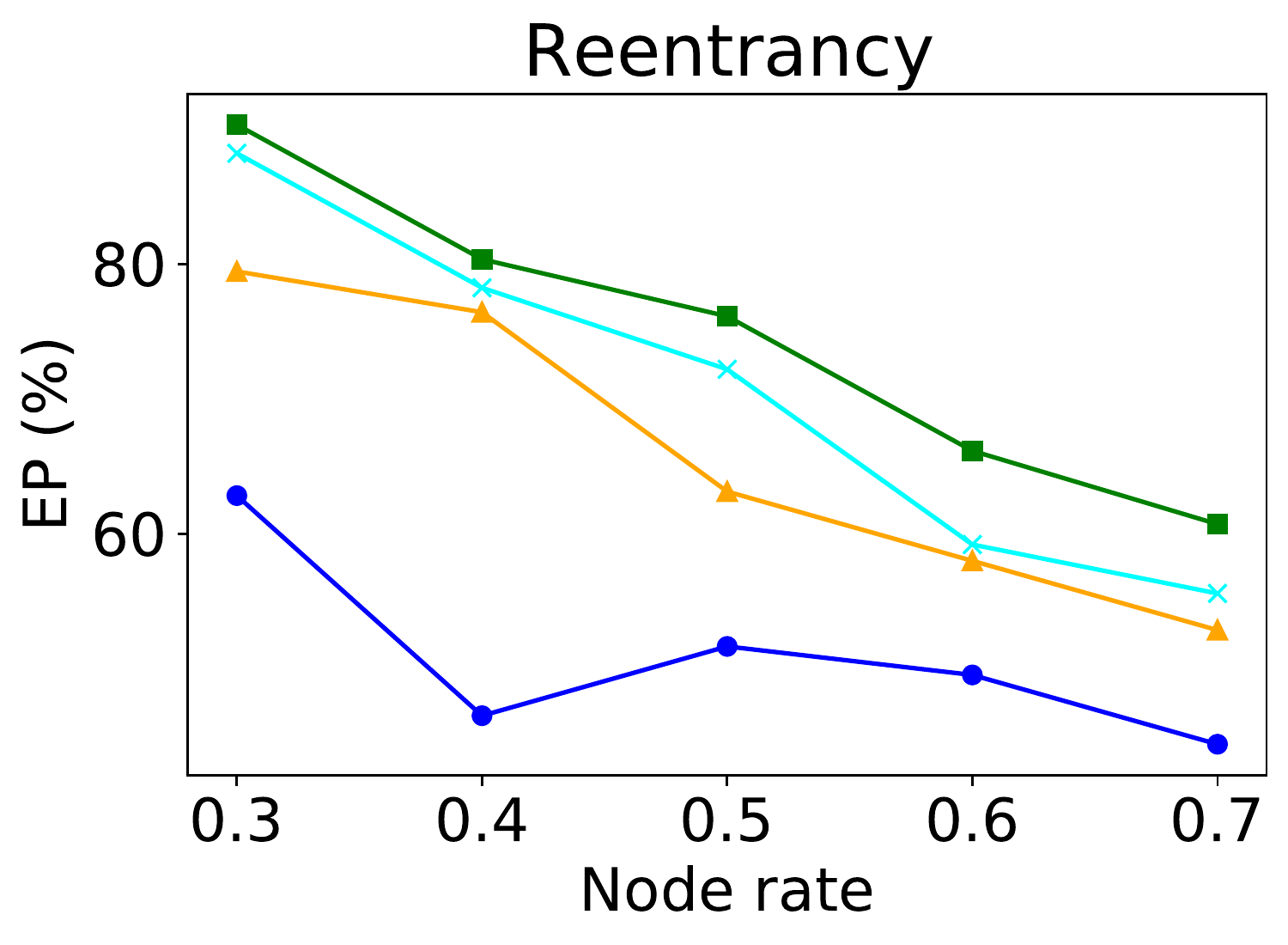}
\end{subfigure}%
\begin{subfigure}{.2\linewidth}
\centering
\includegraphics[width=\linewidth]{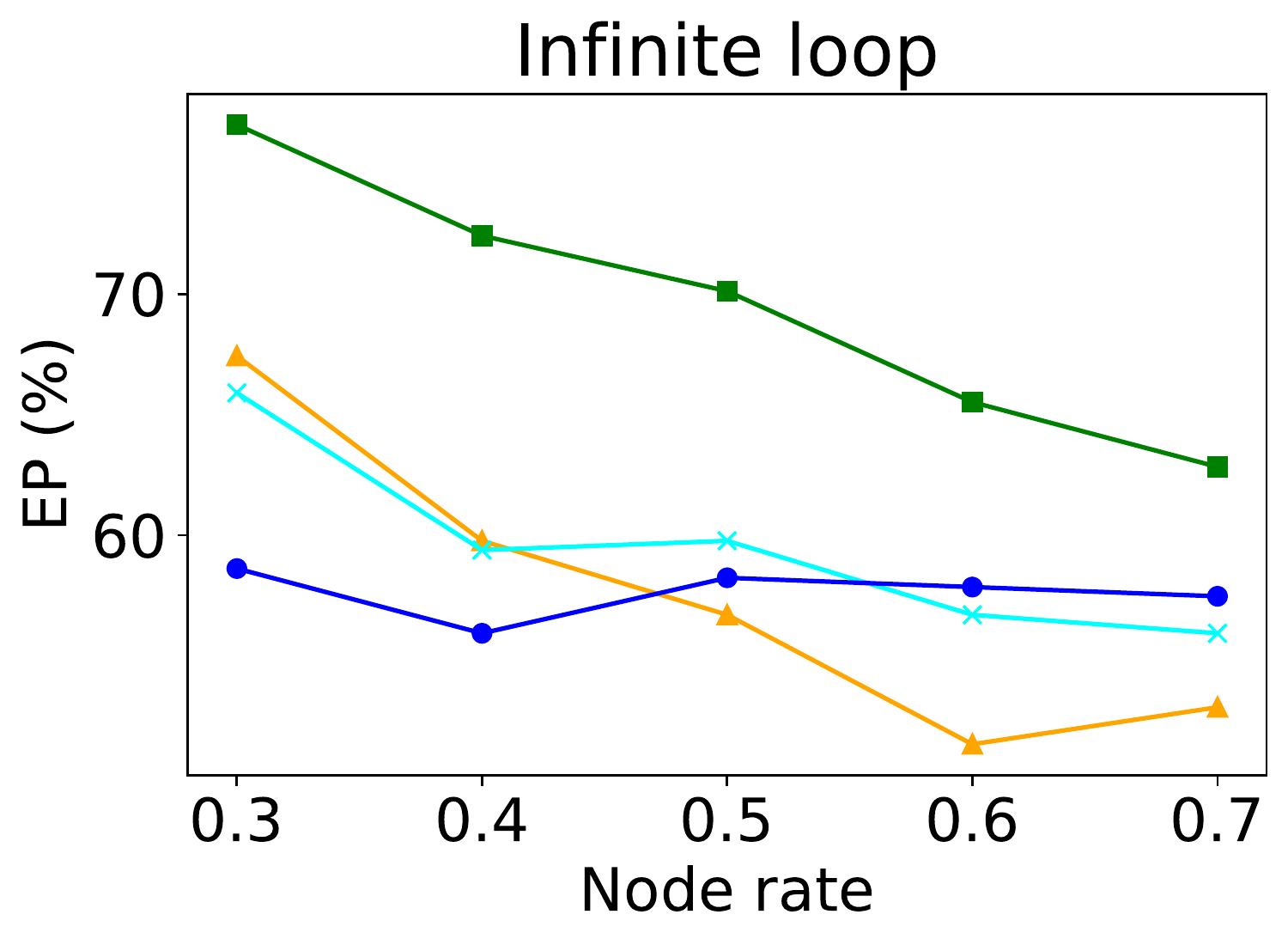}%
\end{subfigure}%
\begin{subfigure}{.2\linewidth}
\centering
\includegraphics[width=\linewidth]{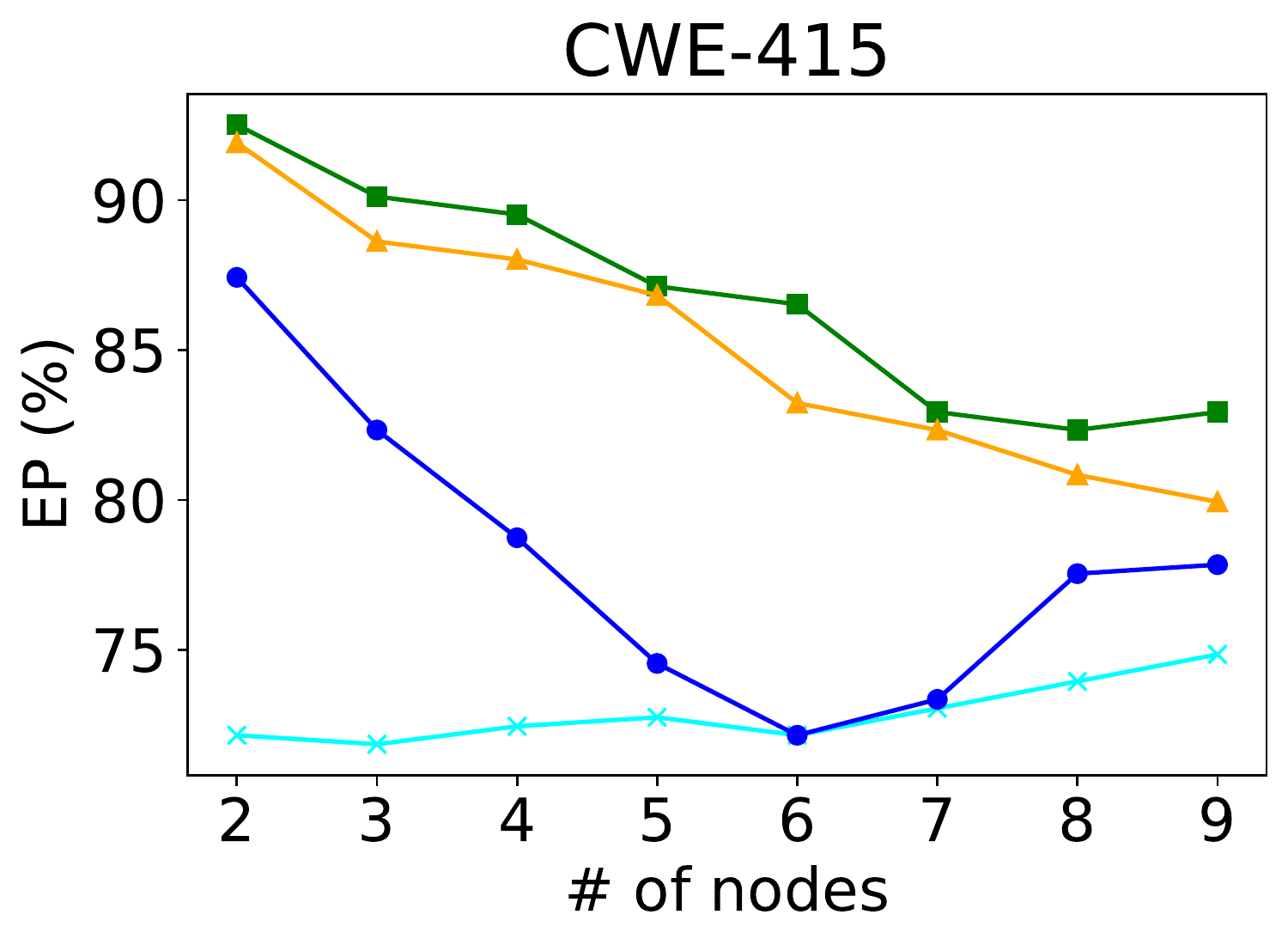}
\end{subfigure}%
\begin{subfigure}{.2\linewidth}
\centering
\includegraphics[width=\linewidth]{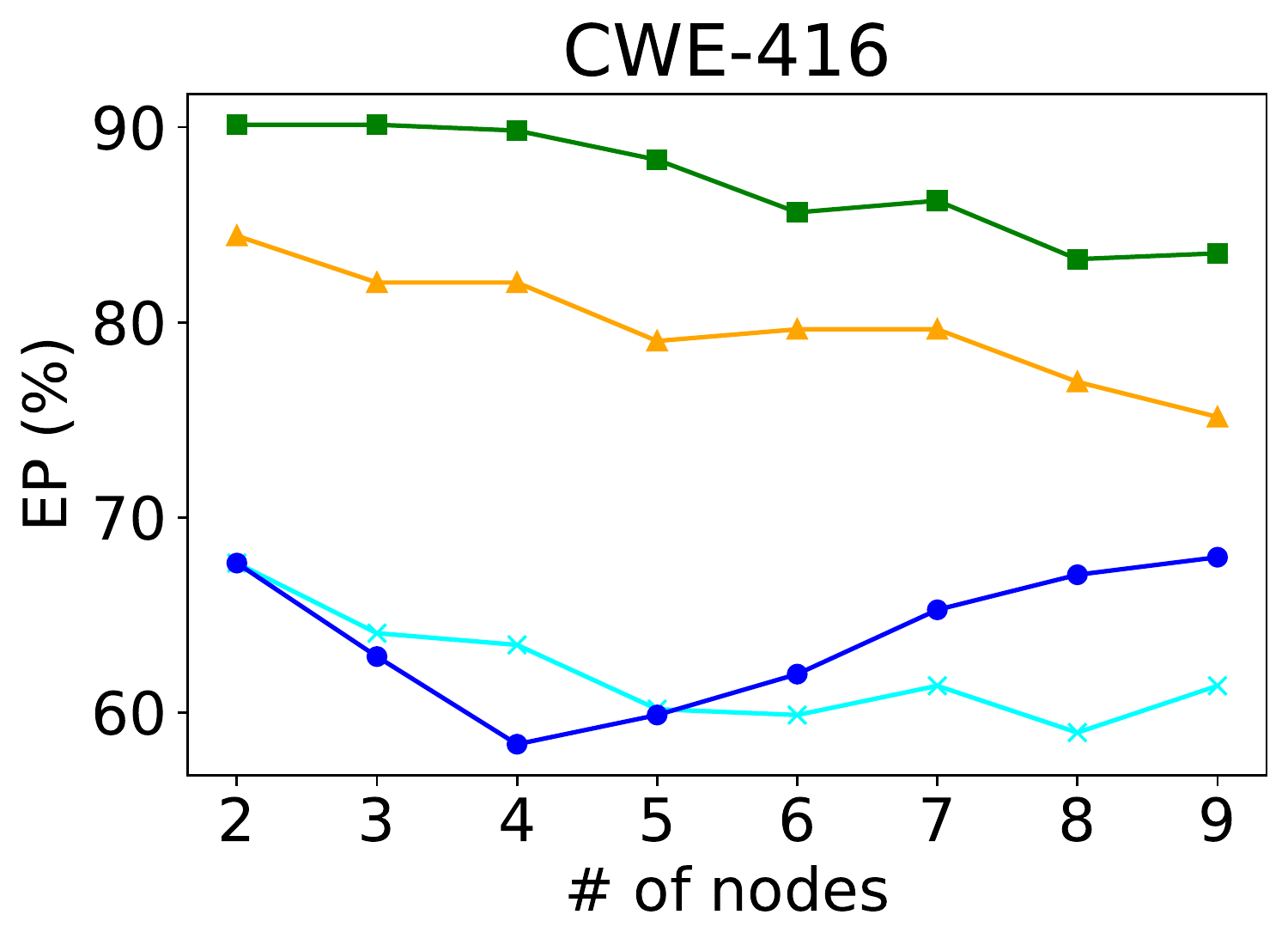}
\end{subfigure}%
\begin{subfigure}{.2\linewidth}
\centering
\includegraphics[width=\linewidth]{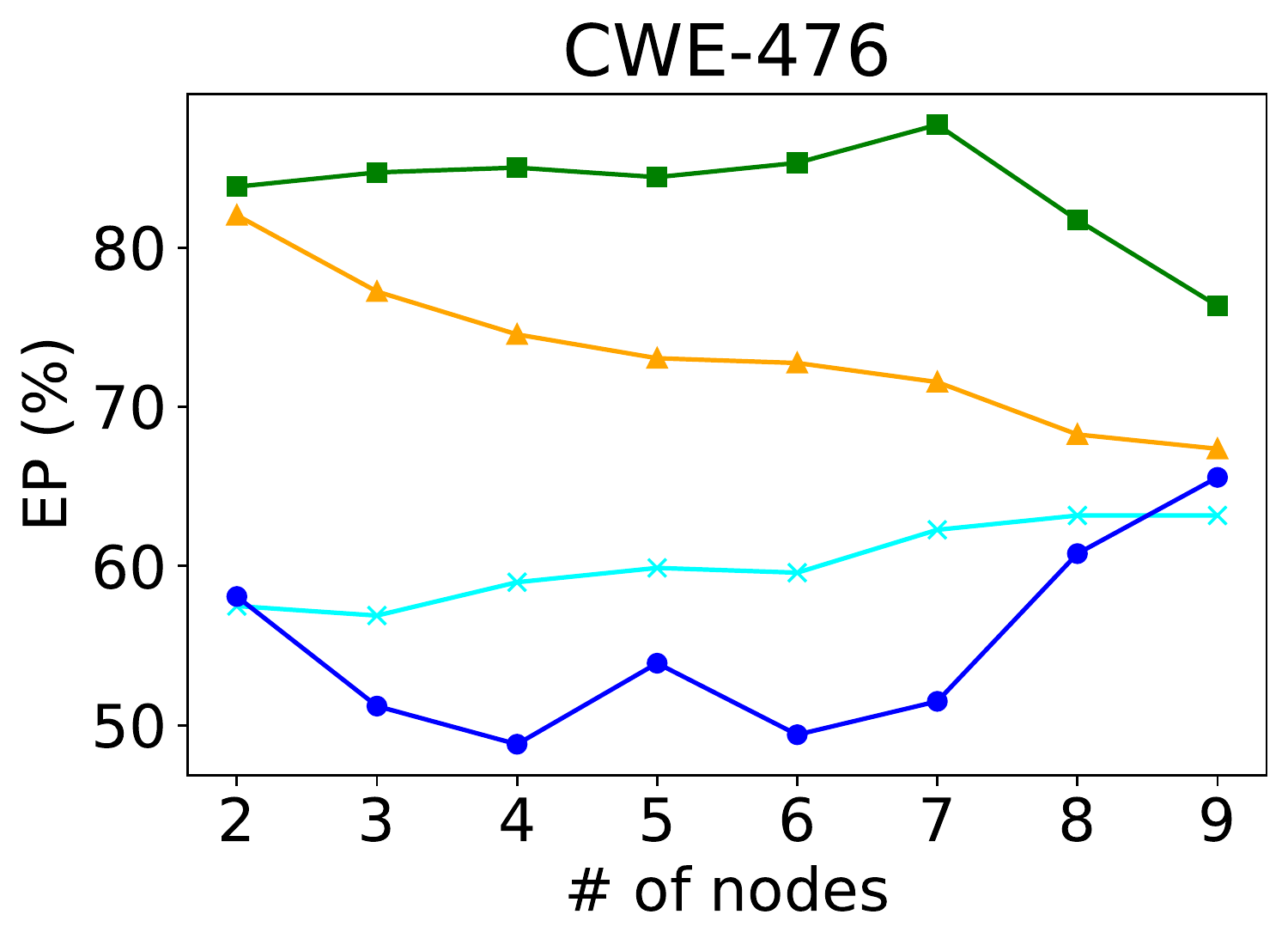}
\end{subfigure}%
\caption{The explanation results for cybersecurity applications. We obtain EP of the remaining subgraphs previously generated. The graph sizes here are for the explained subgraphs.}
\label{fig:ep-}
\end{figure*}

\begin{table*}[t]
\caption{Minimum graph size to retain the original GNN predictions ($Sparsity$).}
\label{tb:result_sparsity}
\centering
\tabcolsep = 0.5cm
\begin{tabular}{lrrrrr}
\toprule
\textbf{Methods} & \textbf{Reentrancy} & \textbf{Infinite loop} & \textbf{CWE-415} & \textbf{CWE-416} & \textbf{CWE-476} \\ 
\midrule
PGM-Explainer & 3.047 & 2.695 & 12.043 & 16.628 & 3.192 \\
GNNExplainer & 2.184 & 2.305 & 9.304 & 14.802 & 2.768 \\
PGExplainer & 2.118 & 2.290 & \textbf{5.928} & 9.407 & 2.838 \\
\textbf{{\name}} & \textbf{2.000} & \textbf{2.015} & 6.406 & \textbf{8.267} & \textbf{1.404} \\
\midrule
Average graph size & 4.939 & 3.695 & 13.029 & 20.047 & 9.838 \\
\bottomrule
\end{tabular}
\end{table*}

\begin{table*}[t]
\caption{Time complexity (seconds).}
\label{tb:result_time}
\centering
\tabcolsep = 0.5cm
\begin{tabular}{lrrrrr}
\toprule
\textbf{Methods} & \textbf{Reentrancy} & \textbf{Infinite loop} & \textbf{CWE-415} & \textbf{CWE-416} & \textbf{CWE-476} \\ 
\midrule
PGM-Explainer & 93.3 & 62.7 & 292.4 & 367.5 & 269.3 \\
GNNExplainer & 37.8 & 35.6 & 92.9 & 94.2 & 91.8 \\
PGExplainer(training) & 0.8(68.3) & 0.6(52.8) & 2.4(83.5) & 3.2(118.8) & 3.0(100.9) \\
\textbf{{\name}} & 52.5 & 37.6 & 99.3 & 103.4 & 98.7 \\
\bottomrule
\end{tabular}
\end{table*}

\textbf{EP of remaining subgraphs.}
Furthermore, we study the EP of remaining subgraphs, which is computed from the rest of top-$R$ or top-$K$ nodes. Therefore, the lower EP represents higher irrelevance of remaining subgraphs. EP is based on the intuition where the remaining input is irrelevant to the prediction since the important factors are identified and removed. We use the same values of top-$R$ and top-$K$ from the evaluation in Figure~\ref{fig:ep+}. 
We show the results for the two applications in Figure~\ref{fig:ep-}, with Table~\ref{tb:ep-} showing the results in the middle from Figure~\ref{fig:ep-}. The EP for {\name} and PGExplainer is overall lower than other explanation methods, meaning the remaining subgraphs are less related to the GNN predictions.
As it is observed in the pair of Figure~\ref{fig:ep+} and Figure~\ref{fig:ep-}, the increase of EP of explained subgraphs does not directly relate to the decrease of EP of remaining subgraphs. 

Our objective is to identify the important subgraphs that retain the original predictions, while the interaction of the remaining nodes can contribute to the prediction as well. GNNs are complex and non-linear models. The important subgraphs are not assembled by all the important nodes individually, but the important node interactions. The remaining subgraphs may contain positive node interactions and important nodes, which are weaker than the explained subgraphs. Thus, the objectives of obtaining the maximum EP of explained subgraphs and the minimum EP of remaining subgraphs are better considered separately, especially for complex models like GNNs.
It is proved that GNNs can be attacked easily by correctly identifying important nodes. The domains of attack and explanation share common techniques, e.g., counterfactual explanation. With the explanation method, the attack can be conducted by removing important nodes or identifying important nodes for an incorrect prediction.

\textbf{Sparsity.}
By default, GNNs are able to make a certain prediction from an empty graph. The default prediction for smart contract vulnerability detection is vulnerable, while the code vulnerability detection is benign. To better differentiate the performance for each explanation method, the $Sparsity$ is only evaluated from graphs with the opposite default predictions. 
We collect the $Sparsity$ from different explanation methods in Table~\ref{tb:result_sparsity}. Overall, {\name} achieves the smallest $Sparsity$, which is consistent with the result of EP of explained subgraphs. For graphs with bigger sizes from code vulnerability detection, it is only fewer than half of the nodes that lead to the final predictions. It indicates the vulnerability does not take a big part of the code, based on the assumption that GNNs make the prediction by correctly capturing the vulnerability factors. From CWE-476, the GNN identifies the significant difference between benign and vulnerable code since it is able to determine the vulnerability averagely within two nodes. The way GNN makes predictions for this dataset is mainly to find the benign factors rather than vulnerable factors. It also implies that the dataset may not be strong or complete enough to cover all the possible coding situations, as GNN only needs to capture the difference between the graphs with different labels.

\begin{table*}[t]
\caption{EP (\%) of explained subgraphs for attribute explanation study. We pick top-$3$ node attributes for smart contract vulnerability detection; top-$5$ for code vulnerability detection.}
\label{tb:result_feat}
\centering
\tabcolsep = 0.5cm
\begin{tabular}{lrrrrr}
\toprule
\textbf{Methods} & \textbf{Reentrancy} & \textbf{Infinite loop} & \textbf{CWE-415} & \textbf{CWE-416} & \textbf{CWE-476} \\ 
\midrule
GNNExplainer & 
74.3 & 64.0 & \textbf{94.3} & \textbf{87.4} & 88.9 \\
\textbf{{\name}} & 
\textbf{92.7} & \textbf{71.6} & \textbf{94.3} & 85.3 & \textbf{98.5} \\
\bottomrule
\end{tabular}
\end{table*}

\begin{table*}[t]
\caption{EP (\%) of explained subgraphs for ablation study. $R=0.5$ for smart contract vulnerability detection; $K=6$ for code vulnerability detection.}
\label{tb:ablation}
\centering
\tabcolsep = 0.5cm
\begin{tabular}{lrrrrr}
\toprule
\textbf{Methods} & \textbf{Reentrancy} & \textbf{Infinite loop} & \textbf{CWE-415} & \textbf{CWE-416} & \textbf{CWE-476} \\ 
\midrule
Edge only & 
83.4 & 72.8 & 82.9 & 74.9 & 80.2 \\
Attribute only & 
67.4 & 72.0 & 83.5 & 81.4 & 95.5 \\
\bottomrule
\end{tabular}
\end{table*}

\textbf{Time complexity.}
Table~\ref{tb:result_time} shows the execution time for every explanation method. We use the same training split from Table~\ref{tb:general_result} for training PGExplainer. 
GNNExplainer overall generates the fastest explanation since it directly and only learns edge masks from each graph (as for graph structure). The extra training cost from PGExplainer takes the majority of the time consumption, while extra mask learning is not needed for explanation. PGM-Explainer spends its running time in node attribute perturbation and calculation. The time consumption is affordable for simple datasets because the graph size is limited and PGM-Explainer provides the accurate explanation, while for complex cybersecurity datasets, more energy is needed for sampling the perturbed dataset. The time complexity of {\name} is closely higher than GNNExplainer due to more time consumption for the nodes and attributes. The time consumption from {\name} is acceptable since {\name} provides a comprehensive and accurate explanation. Large time complexity will be necessary if different explanation methods are combined for a comprehensive explanation.

\subsection{Ablation Study}

\textbf{Attribute explanation study. }
We further evaluate the node attribute explanation of {\name}, as shown in Table~\ref{tb:result_feat}. Generally, the highest EP values are obtained by {\name}. It proves that the node attributes contribute to the prediction differently, so the importance scores should be applied to them individually. 

The results also indicate that only a small number of node attributes are highly important to the prediction. Compared with node explanation, an individual node attribute can contribute more to the prediction than an individual node from the two applications. Intuitively, the attack on node attributes can be easily conducted. Besides, the attack is not as noticeable as the attack on nodes, especially for CWE-476 dataset.

\textbf{Ablation study for node explanation. }
The node importance scores are gathered by the importance scores of message passing, requiring the importance scores for edges and node attributes. Here, we gather the importance scores for nodes by edge explanation only and attribute explanation only, in order to verify the node explanation requires both edge and node attribute explanation. The importance scores from edges only are gathered in the same way as above experiments without considering importance scores from node attributes. The importance scores from node attributes only are gathered from synchronized attribute mask learning. We evaluate the EP of explained subgraphs in Table~\ref{tb:ablation}.

Compared with the results in Table~\ref{tb:ep-}, generally, the node explanation by edge only or attribute only is not as accurate as when they are combined. 
Attribute-only explanation overall obtains lower EP in smart contract vulnerability detection but higher EP in code vulnerability detection. By comparing the difference, the results from Teentrancy indicates the graph structure makes the key contribution to the prediction, while those from CWE-416 and CWE-476 indicate the opposite. Node attributes can take an important role to estimate the importance of each node. For graph structure explanation, especially when it comes to unimportant node removal, it is necessary to have nodes specially explained. 

{\revision
\subsection{Evaluation on Node Classification Task}

Additionally, We study the explanation performance on node classification task. 

\textbf{Background.}
We use the basic Graph Convolutional Network (GCN)~\cite{kipf2017semi} as the node classifier. GCN is a GNN with the following propagation rule for one layer:
\begin{equation}
	H^{(l+1)} = \phi (\Tilde{D}^{-\frac{1}{2}}\Tilde{A}\Tilde{D}^{-\frac{1}{2}}H^{(l)}W^{(l)})
	\text{.}
\end{equation}
Here, $\phi$ is the activation function, $A$ is the adjacency matrix, $\Tilde{A} = A + I$, and $\Tilde{D}_{ii} = \sum_{j}\Tilde{A}_{ij}$. 
For node classification task, fully connected layers are adopted after GCN to compute the classification. 

\begin{table}[t]
    {\revision
	\caption{\revision The specifications of different dataset and the accuracy of the pre-trained models.}
	\label{tb:data}
	\centering
	\tabcolsep = 0.25cm
	\begin{tabular}{lccccc}
		\toprule
		\textbf{Dataset} & $\boldsymbol{\mid \mathcal{V}\mid}$ & $\boldsymbol{\mid \mathcal{E}\mid}$ & $\boldsymbol{\mid Y\mid}$ & $\boldsymbol{\mid \mathcal{X}\mid}$ & \textbf{Accuracy} \\ 
		\midrule
		Cora & 2,708 & 5,429 & 7 & 1,433 & 0.807 \\
		Citeseer & 3,327 & 4,732 & 6 & 3,703 & 0.711 \\
		\bottomrule
	\end{tabular}
	}
\end{table}%

\textbf{Evaluation.}
We use a 2-layer GCN with 64 hidden channels for each layer, and a fully connected layer after GCN for node classification. We adopt ReLU as the activation function. The training and testing split is the public fixed split from ~\cite{yang'16}. Table~\ref{tb:data} shows the information of the dataset we use. 
We use the test split for the explanation. We compare {\name} with GNNExplainer~\cite{GNNEx19}. Here, we extract the top-5 and top-10 nodes for both datasets and evaluate the performance with the metric Essentialness Percentage (EP). 
Since we use 2-layer GCN, the extracted nodes are within 2-hop neighbors.  

\begin{figure}[t]
	\centering
	\includegraphics[width=0.9\linewidth]{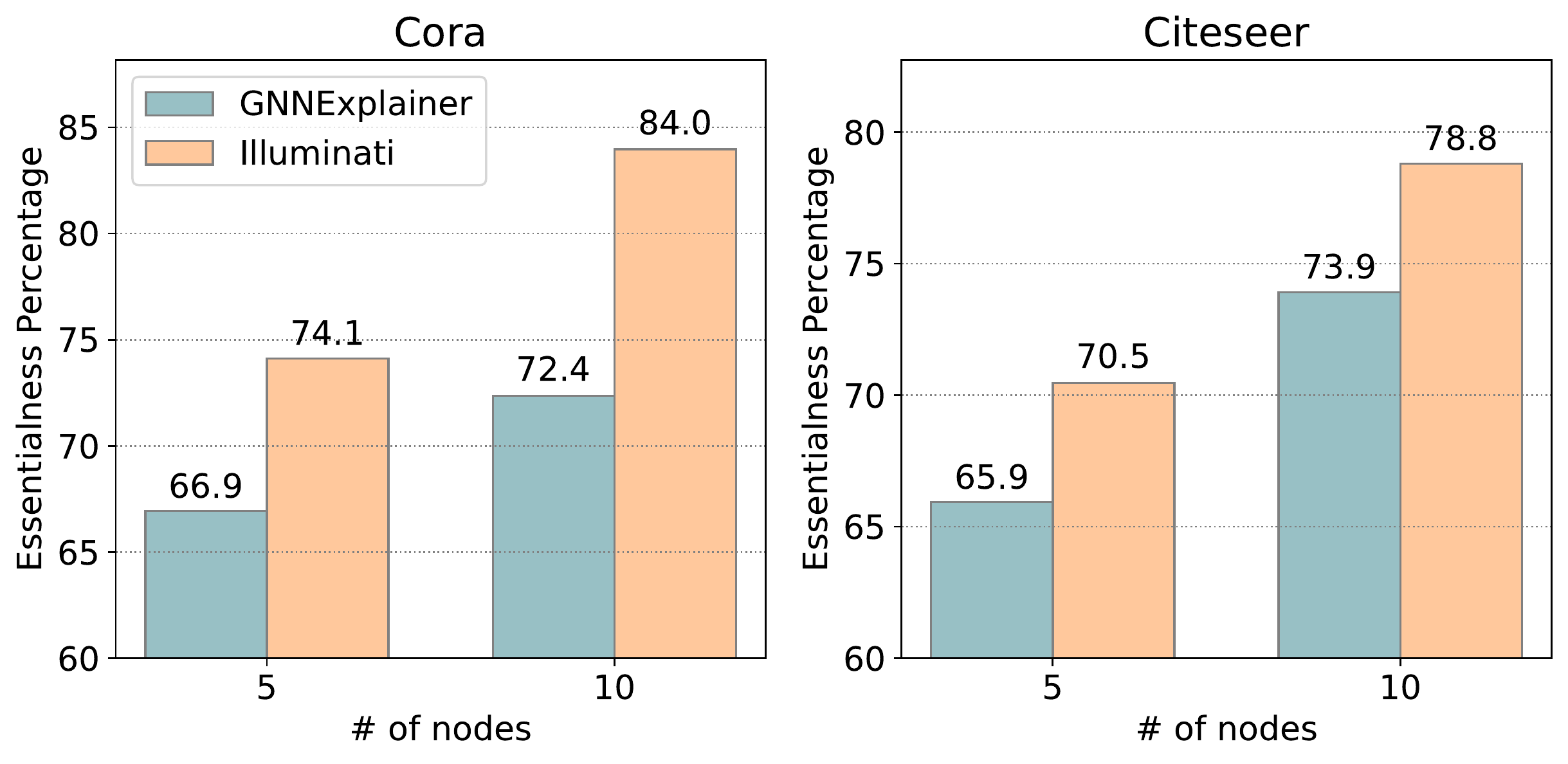}
	\caption{\revision The explanation result of node classification tasks.}
	\label{fig:node_result}
\end{figure}%

As shown in Figure~\ref{fig:node_result}, {\name} obtains 7.1\% EP higher than GNNExplainer on average. From both datasets, {\name} outperforms GNNExplainer distinctly when the number of extracted nodes is small. Such a promising result also proves that it is necessary to jointly consider edges and attributes for node explanation. We believe {\name} will outperform significantly on GNN adaptations and alleviate the limitations of general explanation methods in cybersecurity applications.
}

\section{Case Study}
\label{sec_case_study}

%

In this section, we make two case studies of applying {\name} to real cybersecurity applications, code vulnerability and smart contract vulnerability detection.
In order to obtain straightforward results and comprehensive evaluations, we focus on code vulnerability detection.

\begin{figure*}[t]
	\centering
	\includegraphics[width=.85\linewidth]{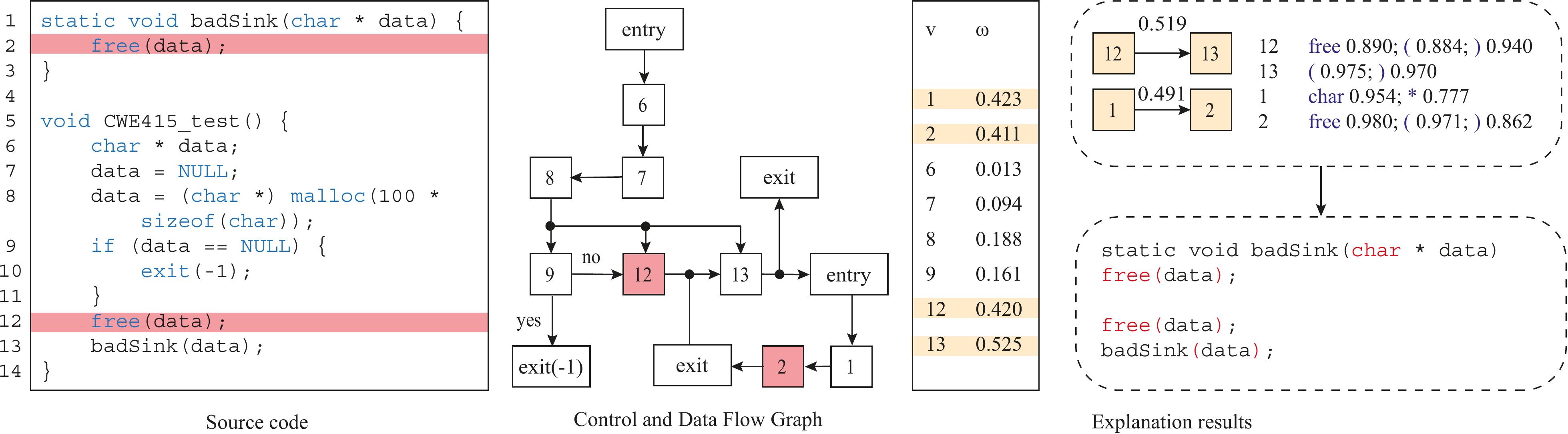}
	\caption{The case study for ``double free''. The reduction in prediction accuracy is 0.005, from the original 1.000.}
	\label{fig:cwe415-164}
\end{figure*}

\begin{figure*}[t]
	\centering
	\includegraphics[width=.85\linewidth]{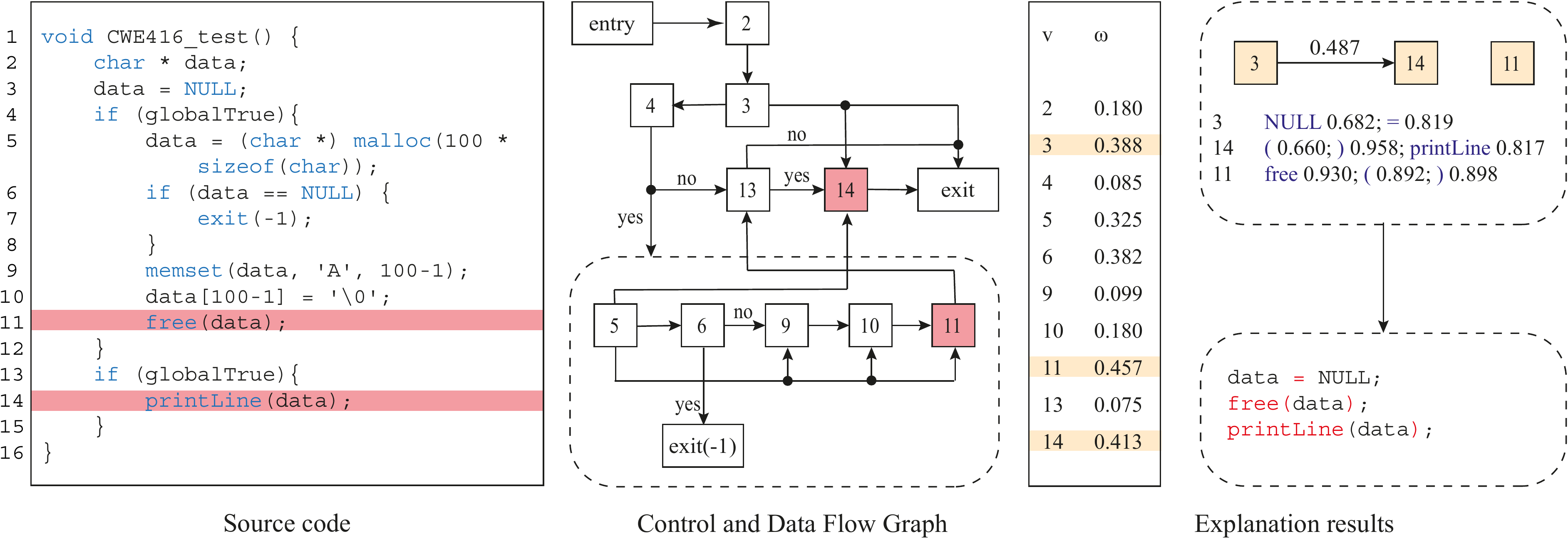}
	\caption{The case study for ``use after free''. The reduction in prediction accuracy is 0.966, from the original 0.980.}
	\label{fig:cwe416-69}
\end{figure*}

\begin{figure*}[t]
	\centering
	\includegraphics[width=.85\linewidth]{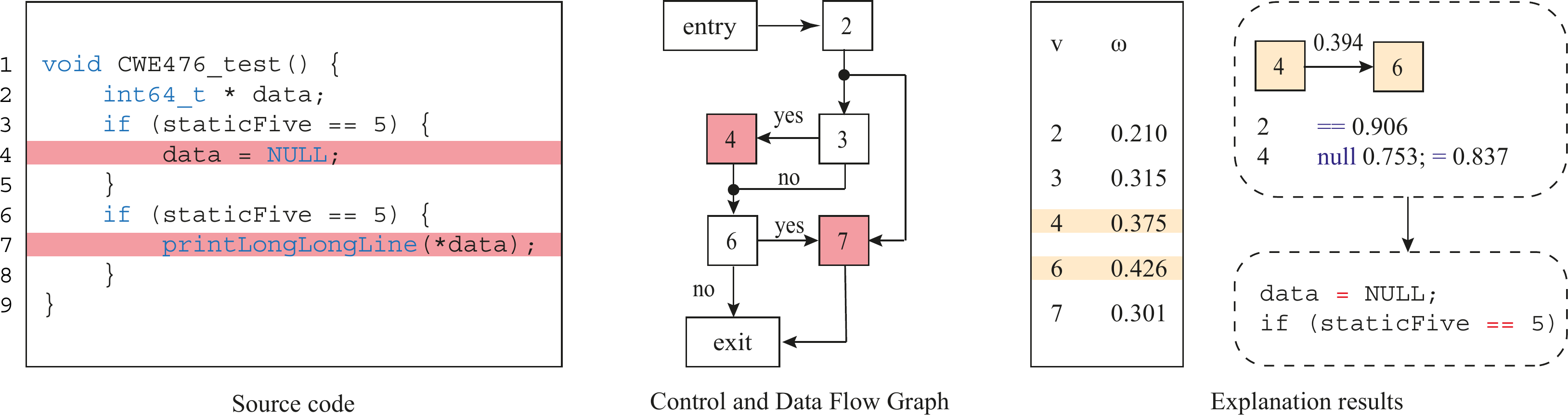}
	\caption{The case study for ``NULL pointer dereference''. The reduction in prediction accuracy is 0.003, from the original 1.000.}
	\label{fig:cwe476-860}
\end{figure*}

\subsection{Case \#1: Code Vulnerability Detection}

\begin{figure*}[t]
	\centering
	\includegraphics[width=.85\textwidth]{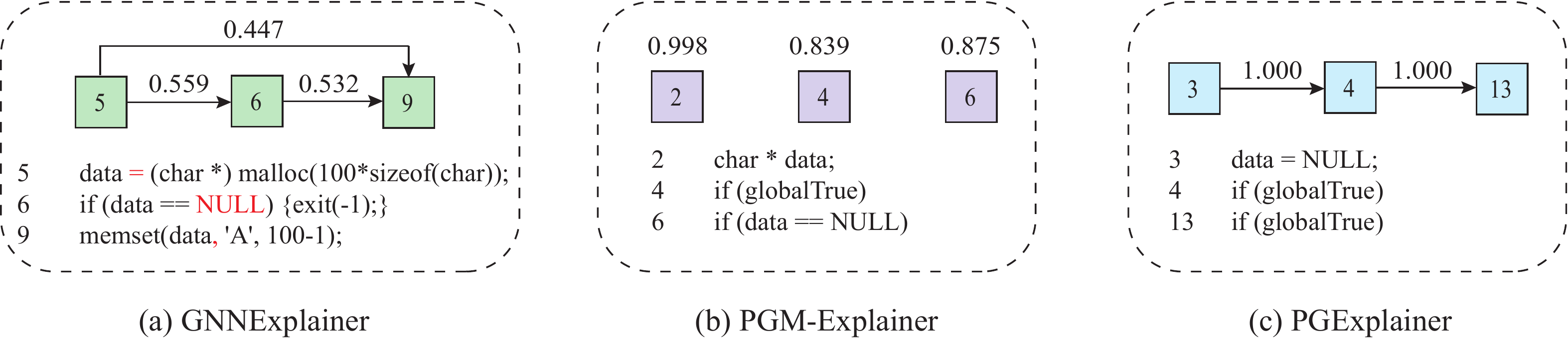}
	\caption{\revision The explanation result from ``use after free'' example. The accuracy reductions are 0.973, 0.980 and 0.977, respectively.}
	\label{fig:case_comparison}
\end{figure*}


\textbf{Background.}
We summarize three steps for code vulnerability detection using GNN models. 
(1) Graph extraction. Code property graphs (CPGs) are generated as the graph representation for source code. A node represents a program construct such as variables, statements, and symbols; an edge contains the direction and relationship information for a pair of nodes such as control flow and data flow. 
(2) Attribute encoding. To better represent the source code and fit the code property graphs to GNNs, node or edge attributes have to be encoded. Node attributes are the most widely used attributes in code vulnerability detection. 
(3) Model learning. This application is conducted as a graph classification task. With the code property graphs and node attributes as input, labels of benignity and vulnerability as targets, the model is learned from a set of datasets. 

In this experiment, we use AFGs as our CPGs. Therefore, a node denotes a statement, an edge contains the direction and relationship information (control flow and data flow) for a pair of nodes. We use Joern~\cite{joern} to extract CDFGs from C/C++ code. We make sure each graph contains 32 nodes. The keywords from each statement are extracted for node attribute encoding. A node attribute indicates whether the statement has the corresponding keyword, e.g., \texttt{char}, \texttt{==} \texttt{*}, so it is encoded to be binary. There are 96 node attributes for each node. We use the model, Devign, as the code vulnerability detector.

\textbf{Evaluating the output of {\name}.}
We measure the reduction in prediction accuracy for each case, which is the probability decrease of the explained subgraph.

The vulnerability in Figure~\ref{fig:cwe415-164} is caused by ``double free''. Different from Figure~\ref{fig:comparison}, the source code here calls a function. The key reasons for vulnerability are the same, while the model considers the function nodes in Figure~\ref{fig:cwe415-164} as the contribution. It is reasonable that the function is the path from line 12 to line 2. The output of {\name} suggests the model's competence and weakness. It successfully captures the vulnerability, but the performance drops down as the source code becomes complex. From our graph generation technique, the functions are not specially identified, which can be opened up and embedded into the major function.

Figure~\ref{fig:cwe416-69} shows an example from the dataset ``use after free''. The output suggests that the model's decision-making is the same as human knowledge. The importance score of the edge indicates the edge is not highly important to the prediction, which may be a potential risk. 

\begin{figure*}[t]
	\centering
	\includegraphics[width=\textwidth]{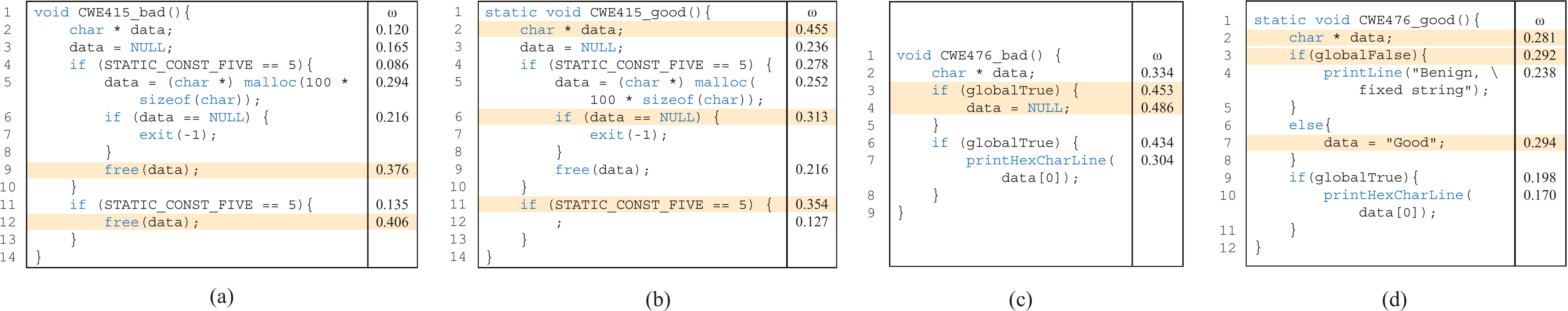}
	\caption{The explanation results of two pairs of mirrored source codes. }
	\label{fig:codes}
\end{figure*}

\begin{figure*}[t]
	\centering
	\includegraphics[width=.85\linewidth]{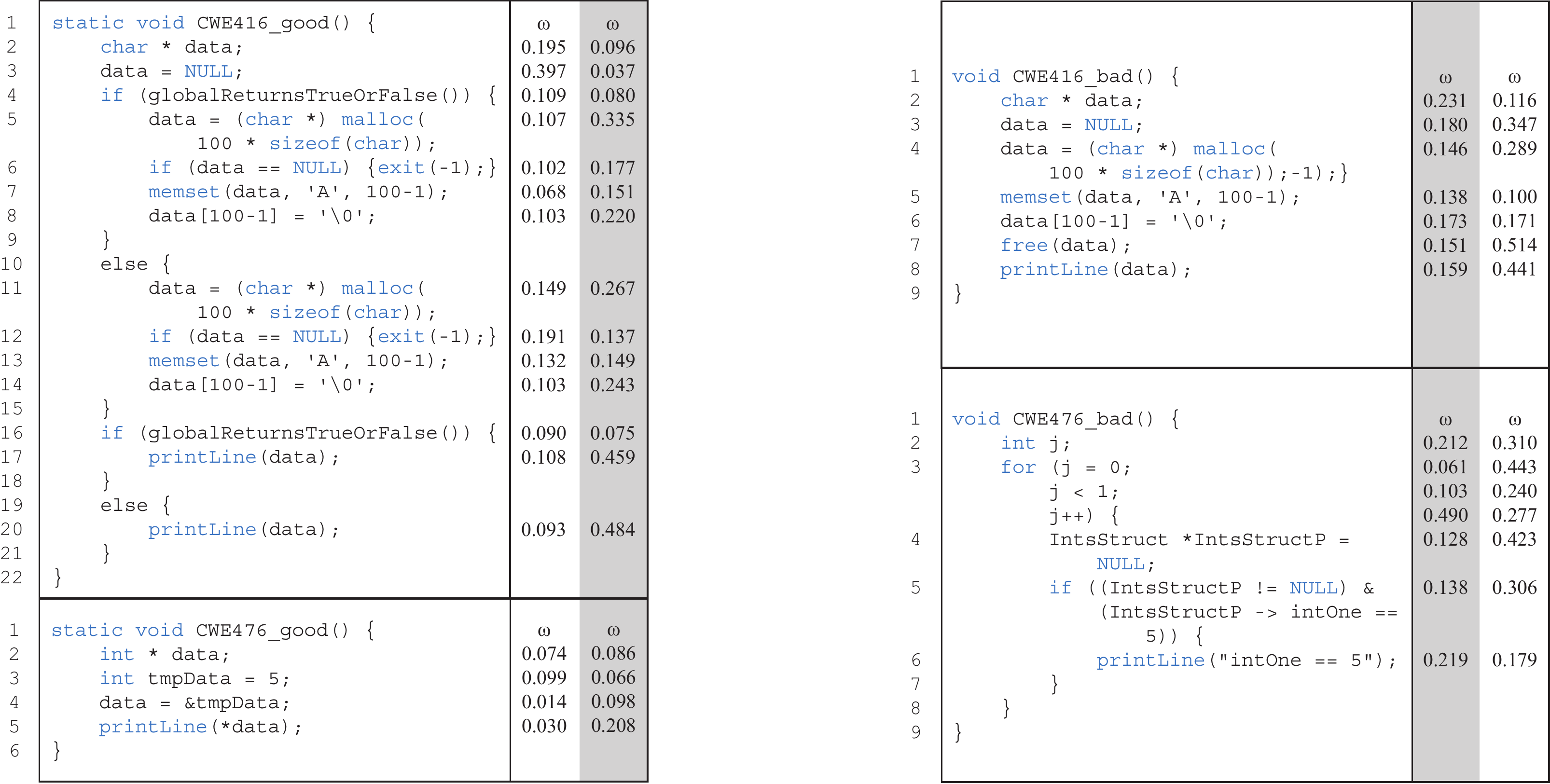}
	\caption{The explanation results of mispredictions. The gray box is the explanation for the mispredictions.}
	\label{fig:mispred}
\end{figure*}

The dereference of \texttt{NULL} pointer leads to the vulnerability in Figure~\ref{fig:cwe476-860}. The explanation results suggest the key reason for the prediction is node 4, where the pointer is assigned as \texttt{NULL}. However, it captures line 6 rather than 7, which is contradictory to human understanding. It is understandable because its mirrored version of benign code contains the symbol \texttt{!=} in \texttt{if} condition (line 6). The explanation suggests the dataset is well learned by the model but less confidence if the model is adopted to real applications.



%


{\revision
Figure~\ref{fig:case_comparison} shows the explanation result from other methods for the same vulnerable code shown in Figure~\ref{fig:cwe416-69}. One can observe that {\name} significantly outperforms other explanation methods by providing a comprehensive explanation for nodes, edges, and attributes. Missing one explanation factor can cause significant difficulty for analysis. The explanation accuracy is also degraded as seen from the reductions in prediction accuracy. PGExplainer, as a global explanation method, may not provide customized results for a single input graph. While the reduction in prediction accuracy is significant, {\name} achieves the lowest reduction and provides human-understandable explanation. As it is observed from Figure~\ref{fig:ep+} and Table~\ref{tb:result_sparsity}, the graph size plays an important role in the prediction. Wrong information from explanation methods may lead to more confusion and the wrong conclusion to the models. With the trust of {\name}, cybersecurity analysts can easily map the output to the source code and understand the model behavior. 

Besides, {\name} alleviates the limitations in graph-specific explanation methods: descriptive accuracy (DA), efficiency, robustness, and stability~\cite{Ganz21}. {\name} greatly improves DA and efficiency as the experiment shows. Specifically in code vulnerability detection, which lines of code contribute to the prediction is important to cybersecurity analysts. Each line is represented as a node in the AFG, which makes it vital to accurately determine the importance of nodes. {\name} accurately identifies the important lines and keywords. By gathering both edge and attribute information for node explanation, {\name} is robust against edge perturbation. Similarly, we believe stability is also preserved.
}

{\revision

\textbf{Using the output of {\name}.}
The explanation methods with high EP should be able to provide accurate information on which part of the code is considered vulnerable by the model. They can identify the vulnerable lines when the model's decision-making matches human knowledge. However, the usage of {\name} is not limited to this. First, {\name} helps cybersecurity analysts pinpoint the model's misbehavior even though the model gives the correct predictions. Second, {\name} helps analysts interpret why mispredictions are made. The developers can identify the pitfalls observed from the recent study~\cite{Arp22}, and take certain actions to troubleshoot and optimize the model based on the output of {\name}.

}



More results of paired code are shown in Figure~\ref{fig:codes}. {\name} detects the important vulnerable factors in Figure~\ref{fig:codes}(a) and (c), benign factors in Figure~\ref{fig:codes}(b) and (d) according to their predicted labels.
As the result shows, the code in Figure~\ref{fig:codes}(a) is vulnerable because of ``double free'', where the model captures the vulnerability and {\name} successfully identifies the vulnerable lines. The explanation for Figure~\ref{fig:codes}(b) shows benign statements from the source code. Combining Figure~\ref{fig:codes}(a) and (b), the explanation suggests that the model makes the classification by detecting vulnerable factors. 
The vulnerability in Figure~\ref{fig:codes}(c) is caused by ``NULL pointer dereference''. Comparing Figure~\ref{fig:codes}(c) and (d), the model detects the vulnerability by the value assignment to the variable, where \texttt{NULL} leads to vulnerability. The model also detects the difference from the conditions. From the dataset, vulnerable functions do not contain a lot of ``false'' conditions. The model fails to identify a key statement, i.e., line 7 in Figure~\ref{fig:codes}(c) because vulnerable and benign code both contain such statements. Therefore, the model for this dataset is vulnerable to attacks and is not trustable even it achieves high accuracy. To alleviate the issues, different conditions should be considered to fill the dataset, and more semantic information can be extracted. 

\begin{figure*}[t]
	\centering
	\includegraphics[width=.85\textwidth]{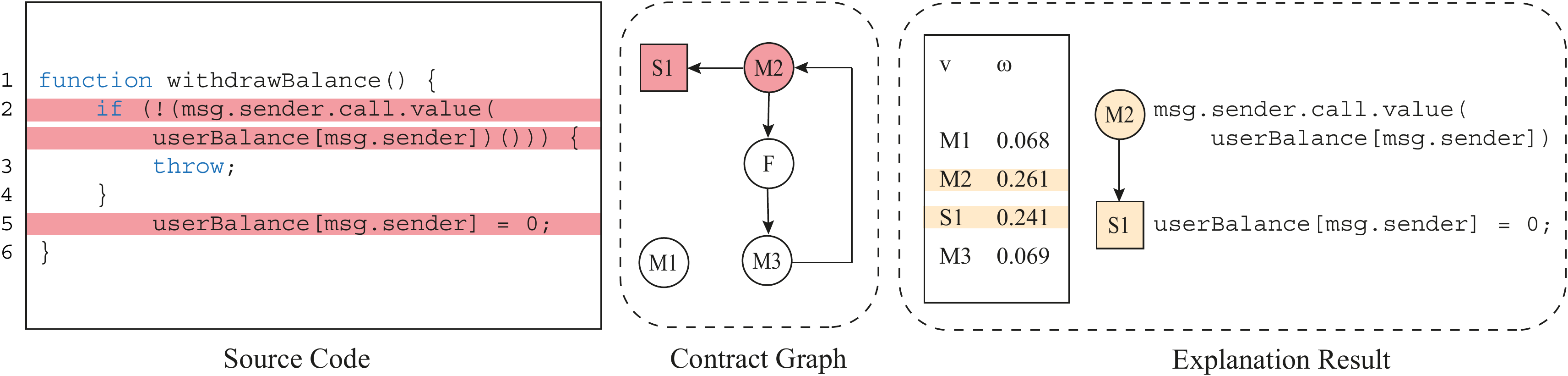}
	\caption{An example of Reentrancy. The reduction in prediction accuracy is 0.332, from the original 0.991.}
	\label{fig:Reentrance_01}
\end{figure*}

\begin{figure*}[t]
	\centering
	\includegraphics[width=.85\textwidth]{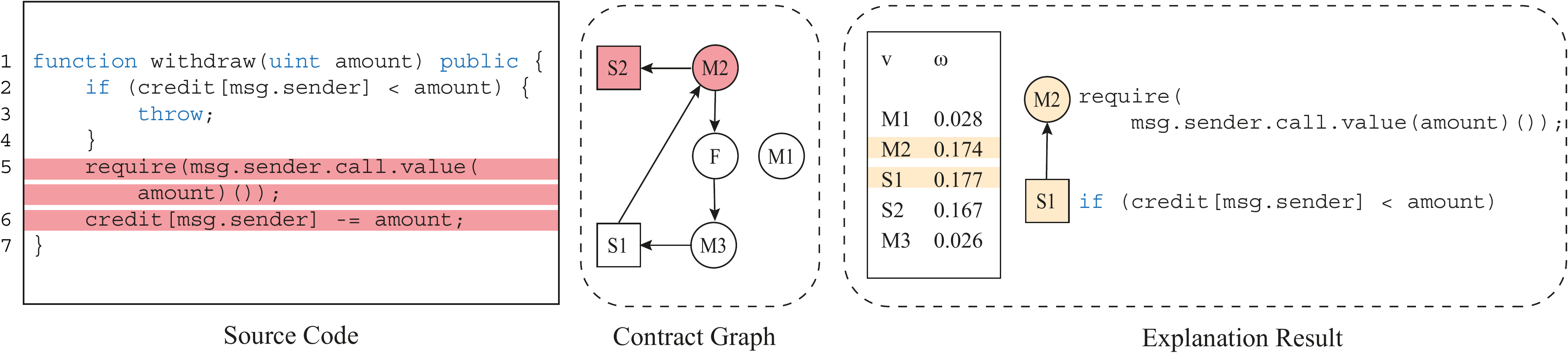}
	\caption{An example of Reentrancy. The reduction in prediction accuracy is 0.186, from the original 0.845.}
	\label{fig:simple_dao}
\end{figure*}

Furthermore, we evaluate cases of mispredictions in Figure~\ref{fig:mispred}, where the gray box is the explanation of GNN predictions (mispredictions). Further explanation for the correct label is also shown in the white box. The labels of the left column are benign, and those on the right are vulnerable. Here we show results from CWE-416 and CWE-476 since the mispredictions from CWE-415 mostly happen to small graphs.
As it can be observed, {\name} suggests GNNs still have captured the important lines for the correct label. The wrong prediction from the left column comes from \texttt{printLine}, which indicates the use of variables in the model's perspective. The model emphasizes the use of variables but fails to determine the variable is not \texttt{NULL}. More different situations should be added into training, e.g., situations of a variable being used by multiple times without being freed in CWE-416. 
The result shows GNNs are able to detect the vulnerability for CWE-416 at the right column. But the benign lines take the lead through the calculation of GNN, as the importance scores in the gray box do not vary largely. The \texttt{for} loop from CWE-476 exist in codes with different labels, so GNNs randomly assign importance of statements in line 3 to different labels. The vulnerability is identified but not strong enough because the use of variables is in an \texttt{if} condition (line 5). \texttt{printLine} usually indicates the use of variables, but here the argument is a string, which is correctly observed as a benign statement.

{\revision
From interpreting the output of {\name}, the pitfalls found in this application includes spurious correlation, the inappropriate performance measures and lab-only evaluation~\cite{Arp22}. Spurious correlation is caused by the artifacts of the dataset. Different coding styles, length of code and logical situations are not completely considered. This can be alleviated with lab-only evaluation by collecting datasets with different cases from the real world. Only evaluating the prediction accuracy may lead to the neglect of the dataset issues. This will give developers the wrong conclusion of the model. Inappropriate performance measures are addressed by strong explanation methods such as {\name}. The developers can interpret the explanation output for the decision-making and the potential risks of the model.
The output of {\name} suggests several internal drawbacks of the models as well, e.g., the model does not learn the semantic meaning. The models we use do not make full use of the source code information. Without enough semantic information of the statements and the type of edges, it prevents the model from making the correct prediction in Figure~\ref{fig:mispred}. The developers can build a solid strategy to improve the model with the output of {\name}.
}

\subsection{Case \#2: Smart Contract Vulnerability Detection}

We consider cases from Reentrancy dataset, as the contract graphs from vulnerable source code contain enough nodes for the case study. 
The contract graph is constructed according to the work of Zhuang \textit{et al.}~\cite{ijcai2020-454}. The nodes in a contract graph are categorized into major nodes, secondary nodes, and fallback nodes. The major nodes represent important functions, the secondary nodes represent model critical variables and the fallback nodes simulate the fallback function. The edges indicate the relationship between nodes, where the edge attributes are only used for graph construction, not in DR-GCN. The node attributes are derived from the types of functions, operations of variables, etc.
Figure~\ref{fig:Reentrance_01} and \ref{fig:simple_dao} show case studies from Reentrancy dataset. The node M1 is the function that calls \texttt{withdraw} function, M2 is the built-in \texttt{call.value} function and M3 is the \texttt{withdraw} function, all of which are major nodes.

The vulnerability in Figure~\ref{fig:Reentrance_01} comes from the value being assigned (line 5) after checking if ether sending (line 2) goes through. From the explanation result, the GNN model successfully identifies the location of vulnerability.

With the same vulnerability in Figure~\ref{fig:simple_dao}, however, the GNN captures the factors leading to the right prediction rather than the vulnerable statements. {\revision From the code, the transaction (line 5) is after the \texttt{if} statement (line 2). So the model predicts the function as vulnerable. The explanation result shows the two key statements for the prediction. But they are not exactly the ground truth causing the vulnerability, so the decision-making of the model is still confusing to users.} To address the issue, we show the mirrored benign code as follows. 

\begin{lstlisting}[basicstyle=\small]
function withdraw(uint amount) public {
    if (credit[msg.sender] >= amount) {
        credit[msg.sender] -= amount;
        require(msg.sender.call.value(
            amount)());
    }
} 
\end{lstlisting}%


From its mirrored benign code, the value assignment and ether sending is under \texttt{if} condition. In the \texttt{if} condition, the value is assigned first, then the \texttt{call.value} function is called. Accordingly, the path in the corresponding contract graph would be S1 $\,\to\,$ S2 $\,\to\,$ M2. Here, S1 does not directly connect with M2, which causes different node representations from the code in Figure~\ref{fig:simple_dao} and they are learned by the GNN model. Thus, a potential problem from the dataset is identified.

{\revision
A common pitfall from the training datasets in the two applications is spurious correlation, specifically the lack of various real-world coding situations. The models may not make the correct predictions in different dataset because the output of {\name} suggests the models have learned some artifacts rather than the real difference between vulnerability and benignity. 
The edge type is also neglected in this application. How developers utilize the output of {\name} and improve the model is similar to code vulnerability detection.
}





\section{Related Work}
\label{sec_related_work}

\textbf{Graph neural networks. }In recent years, there have been a great number of evolutions in GNNs. 
Scarselli \textit{et al.}~\cite{gnn} firstly introduced GNN as a neural network model, extending the traditional neural network for graph data processing. Bruna \textit{et al.}~\cite{bruna2014} extended the convolutional methods for graph structure by analyzing the constructions of deep neural networks on graphs. Defferrard \textit{et al.}~\cite{defferrard2016} proposed the extension of CNNs to graphs using Chebyshev polynomials. GCN identified that the simplifications can be used in the previous work and presented fast approximate convolutions on graphs.
Plenty of the GNN models, including GCN~\cite{kipf2017semi}, GraphSAGE\cite{hamilton2017inductive}, and GAT~\cite{velickovic2018graph}, generate node representations iteratively by aggregating and updating the attributes from the neighbor nodes. 
The node representations, then are used in different tasks like node classification~\cite{kipf2017semi,WuSGC19}, link prediction~\cite{mlink_aaai20,zhang2018link}, and graph classification~\cite{Errica2020A,magcnn_aaai20}.

\textbf{Deep learning explanation. }The generic purpose of an explanation method is to determine the decision-making by a complex deep learning model. The two major classes of an explanation method are black-box based~\cite{lime,lemna} and white-box based~\cite{ancona2018towards,oramas2018visual}. 
Methods with various techniques are proposed to uncover the behaviors of deep learning models. LIME~\cite{lime} and work from paper~\cite{zintgraf2017visualizing} treat the whole deep learning model as a blackbox. The model decision is explained by directly identifying the important factors from the input. Methods such as LRP~\cite{bach-plos15} and DeepLIFT~\cite{deeplift} decomposes the output backward through model layers and explain the contribution of neurons. Rather than providing a post-hoc explanation for deep learning models, CapsNet~\cite{capsnet} is built as a DNN model with the embedded design of explainability.
Some explanation methods work on specific models, e.g., CNN~\cite{Zhang_2018_CVPR} and RNNs~\cite{bradbury2016quasi}.

\textbf{GNN explanation.}
GNNExplainer~\cite{GNNEx19}, as the pioneering explanation method directly targeting on GNNs, provides edge and node attribute explanations by learning the corresponding masks, which represent the importance scores. PGExplainer~\cite{luo2020parameterized} provides an inductive edge explanation method working on a set of graphs, by learning edge masks with a multi-layer neural network. GraphMask~\cite{schlichtkrull2021interpreting}, however, learns the edge masks for each layer of GNNs and predicts whether an edge can be dropped while retaining the prediction. Differently, PGM-Explainer~\cite{pgmexplainer} identifies important nodes by random node attribute perturbation and a probabilistic graphical model. SubgraphX~\cite{subgraphx_icml21} explains graph in node-assembled subgraph level by Monte Carlo tree search with Shapley value as the scoring function. Different explanations for GNNs have recently been explored. CF-GNNexplainer~\cite{cf_gnnexplainer} targets on counterfactual explanations by learning a binary perturbation matrix that sparsifies the input adjacency matrix. 
With the evolution of GNN explanation methods, a recent survey~\cite{yuan2021explainability} categorized graph explanation methods into two major levels --- instance-level and model-level. The aforementioned methods belong to instance-level, which provide explanations for specific inputs. Model-level methods generate a typical graph pattern that explains how the prediction is made. XGNN~\cite{Yuan_2020} directly explains a GNN model by graph generation, using a reinforcement learning method. If trained by multiple graphs, PGExplainer is able to provide model-level explanation. 

\section{Discussion}
\label{sec_discussion}


Our method can be adjusted to different cybersecurity applications using GNNs since it is comprehensive and the importance scores are learned from the feedback of GNNs. The design is based on the common architecture of GNNs without requiring prior knowledge. The experiment further proves that {\name} improves the performance in both graph classification and node classification. 

In this paper, we mainly focus on node attributes as for attribute explanation, while it can be adjusted to different attributes. As the importance scores for edges and attributes are learned, node importance scores are able to be obtained. Several applications including code vulnerability detection construct graphs with edge attributes, but the attributes are not learned by GNNs. Edge attributes, as edge labels in many applications, can be learned and utilized by relational models. Then an edge is denoted as $(i, j, r)$, where $r$ indicates the relationship of the edge. There will be sets of edge lists categorized by the relationships. 

The explainability of GNNs is not as well-explored as other traditional deep learning models. Besides understanding the contributive factors to the prediction, there is a significant space to fill in, e.g., global explanation and causal explanation.
It is observed from the EP of the remaining subgraphs that these subgraphs still contribute to the prediction. Different types of explanations are needed for cybersecurity applications. 
{\name} can easily be adjusted for counterfactual explanations by adopting CF-GNNExplainer~\cite{cf_gnnexplainer}. Due to the similarity between explanation and attack, there is work~\cite{xujing2021} to conduct backdoor attacks against GNNs with explanation methods. {\name} can also be utilized for attack and defense. 




\section{Conclusion}
\label{sec_conclusion}

In this paper, we propose {\name}, an explanation method that provides a comprehensive explanation for GNNs. By learning the importance scores for both graph structure and node attributes, {\name} is able to accurately explain the prediction contribution from nodes, edges, and attributes. 
We apply {\name} to two cybersecurity applications. Our experiments show {\name} achieves high explanation fidelity. 
We also demonstrate the practical usage of {\name} in cybersecurity applications. 

\bibliographystyle{IEEEtran}
\bibliography{IEEEabrv,ref}

\end{document}